\documentclass{article}

\usepackage{microtype}
\usepackage{graphicx}
\usepackage{tabularx,ragged2e,booktabs,caption}
\usepackage{esvect}
\usepackage{booktabs} 
\usepackage{amsmath, amssymb}
\usepackage{bm}
\usepackage{amsthm}
\DeclareMathOperator*{\argmax}{arg\,max}

\newcommand{\Corr}{\mathrm{Corr}}
\usepackage{multirow}
\usepackage{diagbox}
\usepackage{mathtools}
\usepackage{graphicx}
\usepackage{subfig}
\usepackage{wrapfig}
\usepackage{xcolor}
\usepackage{algorithm}
\usepackage{algorithmic}
\usepackage{wrapfig}

\newcommand{\Sigmoid}{\mathop{\mathrm{Sigmoid}}}
\newcommand{\Bernoulli}{\mathop{\mathrm{Bern}}}
\newcommand{\BinConcrete}{\mathop{\mathrm{BinConcrete}}}
\newcommand{\Uniform}{\mathop{\mathrm{Uniform}}}
\newcommand{\Beta}{\mathop{\mathrm{Beta}}}

\usepackage{textcomp}

\usepackage{ragged2e}
\newcommand{\quotes}[1]{``#1''}
\usepackage[nonatbib,preprint]{neurips_2019}

\usepackage[utf8]{inputenc} 
\usepackage[T1]{fontenc}    
\usepackage{hyperref}       
\usepackage{url}            
\usepackage{booktabs}       
\usepackage{amsfonts}       
\usepackage{nicefrac}       
\usepackage{microtype}      
\title{Lifelong Bayesian Optimization}

\author{
Yao Zhang \\
University of Cambridge\\
\texttt{yz555@cam.ac.uk}
\And
James Jordon\\
University of Oxford \\
\texttt{james.jordon@wolfson.ox.ac.uk}
\AND
Ahmed M. Alaa \\
University of California, Los Angeles \\
\texttt{ahmedmalaa@ucla.edu}
\And
Mihaela van der Schaar \\
University of Cambridge \\
\texttt{mv472@cam.ac.uk}
}

\begin{document}

\maketitle

\begin{abstract}
	Automatic Machine Learning (Auto-ML) systems tackle the problem of automating the design of prediction models or pipelines for data science. In this paper, we present Lifelong Bayesian Optimization (\textsc{LBO}), an online, multitask Bayesian optimization (\textsc{BO}) algorithm designed to solve the problem of model selection for datasets arriving and evolving over time. To be suitable for \quotes{lifelong} Bayesian optimization, an algorithm needs to scale with the ever increasing number of acquisitions and should be able to leverage past optimizations in learning the current best model.  In \textsc{LBO}, we exploit the correlation between black-box functions by using components of previously learned functions to speed up the learning process for newly arriving datasets. Experiments on real and synthetic data show that LBO outperforms standard BO algorithms applied repeatedly on the data. 
\end{abstract}

\section{Introduction}
Designing machine learning (ML) models or pipelines is often a tedious job, requiring experience with, and a deep understanding of, both the methods and the data. For this reason, Auto-ML methods are becoming ever more necessary to allow for broader adoption of machine learning in real-world applications, allowing non-experts to use machine learning methods easily \quotes{off-the-shelf}. The problem of pipeline design is a one of model selection and hyper-parameter optimization (which is itself a type of model selection). To perform this model selection, the mapping from model to performance on a given dataset is treated as a black-box function that needs to be optimized. Several AutoML frameworks \cite{KotthoffTHHL17,FeurerKESBH15,OlsonM16,AlaaS18a} have been proposed for the black-box optimization.

In this paper, we address a general problem of (related) datasets arriving over time, in which we want to perform a model selection for each dataset as it arrives. This problem presents itself in many settings. For example, in medicine, data collection is an ongoing process; every hospital visit a patient makes generates new data, and the models we wish to use on the data need to be optimized to perform well on the most recent patient population. Moreover, hospital practices might change, thereby creating a potential shift in the distribution or structure of the data{\textemdash}some features may stop being measured and new ones introduced because of a technological or medical advancement. It is key that past data not simply be discarded and a new optimization run every time new data arrives, but instead past data should be leveraged to guide the optimization of each new pipeline. 

We cast model selection as a black-box function optimization problem. In particular, we assume there is a sequence of {\em related} black-box functions to be optimized (corresponding to a sequence of related datasets). Bayesian Optimization (BO) \cite{SnoekLA12} aims to find an input optimization $\bm{x}_{*}\in \argmax_{\bm{x}\in \mathbb{X}}f(\bm{x})$, corresponding to an optimal model and hyperparameter configuration for the dataset being modelled by $f$. In this setting a sequence of datasets arrives over time, hence we have a {\em sequence} of black-box functions to optimize, $f_1, ..., f_T$. If the dataset at time $T$ is similar to some previous datasets, their optimal hyperparameter configuration will be similar, or the corresponding black-box functions will be correlated. In the literature, Meta Learning and Multitask Bayesian Optimization are two classes of methods to speed up the hyperparameter optimization process for new datasets using the information from past datasets. Our method \textsc{LBO} is closely related to the latter.

\textbf{Meta Learning.} Using additional meta-features to identify which past datasets are likely to be similar, Meta Learning warm starts a BO optimizer with the hyperparameter configurations that are optimal for the most similar previous datasets. There are three different kinds of meta-features: PCA, Statistical and Landmarking meta-features \cite{bardenet2013collaborative,fulkerson1995machine,pfahringer2000meta}. In the space of meta-features, algorithms and distance metrics are proposed to measure the distance between datasets \cite{FeurerKESBH15,bardenet2013collaborative,shahriari2016taking,reif2012meta}. Instead of using meta-features, our method \textsc{LBO} learns the correlation between black-box functions from the acquisition data.

\textbf{Multitask Bayesian Optimization.} In Neural Network based BO, the acquisition function is constructed by performing a Bayesian linear regression in the output layer \cite{SnoekRSKSSPPA15}. However, this approach gives rise to two technical challenges: (1) Marginal Likelihood Optimization requires inverting a $D^{*}\times D^{*}$ matrix in every gradient update where $D^{*}$ is the size of the last hidden layer; (2) Poor variance estimate due to not measure the uncertainty using the posterior distribution of all the neural network parameters. The work \cite{perrone2018scalable} proposes an efficient training algorithm in marginal likelihood maximization. The work \cite{springenberg2016bayesian} uses Bayesian Neural Network \cite{neal2012bayesian} to improve the variance estimate by approximating the posterior distribution of all the neural network parameters using Stochastic HMC \cite{ chen2014stochastic}. Both works \cite{perrone2018scalable, springenberg2016bayesian} assume all the tasks share one single neural network. Assuming there exists a shared model among the tasks, Multitask Learning (MTL) can improve the model generalization with increasing total sample size over all the tasks \cite{maurer2016benefit}. Multitask BO is more sample efficient in optimizing $f_T$ after we have obtained sufficient acquisition data of $f_1, ..., f_{T-1}$ to learn the shared model. However, when $f_T$ is different from the previous black-box functions, enforcing them to share the model may mislead the estimate of $f_T$ when there are insufficient samples of $f_T$. In \textsc{LBO}, the black-box function learns not only the parameters to be shared but also which datasets should share these parameters, thus retaining the benefits of MTL. In GP based BO, the existing multitask BO \cite{SwerskySA13,KleinFBHH17,poloczek2017multi} or Bandit algorithms \cite{Kirthevasan16,Jialin2018} only consider using a fixed number of latent GPs to model the black-box functions. Fixing the number of tasks requires rerunning the Multitask GPs every time a dataset arrives. The complexity of exact inference of Multitask GPs is $O\big((T\times N)^{3}\big)$ where $N$ is the total number of acquisitions and $T$ is the number of black-box functions. Sparse GP approximation \cite{nguyen2014collaborative} can reduce the complexity to $O\big(T\times K^{3}\big)$ where $K$ is the number of inducing points used to approximate each latent Gaussian process. However, the approximation method using inducing points \cite{hensman2013gaussian} is usually only shown to be useful when there are abundant observations in a low-dimensional space. When the number of inducing points is much smaller than the number of observations, the predictive performance deteriorates severely in practice \cite{wilson2015thoughts}. Sparse GPs for single task BO is attempted in \cite{mcintire2016sparse} where the experiment only shows the results within 80 function evaluations. In \textsc{LBO}, we use neural networks to model the black-box functions, without assuming the data is abundant for a sequence of black-box functions.

\textbf{Contribution.} In \textsc{LBO}, we go beyond restricting the black-box functions to share one single neural network or a fixed number of latent GPs. Inspired by Indian Buffet Process (IBP) \cite{Griffiths11}, as datasets arrive over time, we treat the black-box function on each dataset as a new customer arriving in a restaurant; we apply IBP to generate dishes (neural networks) to approximate the new black-box function. \textsc{LBO} learns a suitable number of neural networks to span the black-box functions such that the correlated functions can share information, and the modelling complexity for each function is restricted to ensure a good variance estimate in BO. We develop a simple and efficient variational algorithm to optimize our model.  We demonstrate that \textsc{LBO} improves the computation time of model selection on three datasets collected from United Network for Organ Sharing (UNOS) registry.

\section{Lifelong Bayesian Optimization}
Let $\mathbb{X}$ be some input space\footnote{Typically this will be the product of all hyperparameter spaces for the models being considered along with an additional dimension that will be used to indicate the model.}. Let $f : \mathbb{X} \to \mathbb{R}$ denote some black-box function and $a_f : \mathbb{X} \to \mathbb{R}$ denote the acquisition function that quantifies the utility of evaluating $f$ at $\bm{x}$. In this paper, we extend this formulation by assuming that we have datasets $\mathcal{D}_{t}$ arriving sequentially for time steps $t\in\mathbb{Z}^{+}$. The datasets may overlap if the user chooses to combine the newly arriving data with the old ones. In this case, we let $\mathcal{D}_{t}$ denote the updated dataset at time $t$. Assume there is a corresponding black-box function $f_t: \mathbb{X} \to \mathbb{R}$, each mapping from model-and-hyperparameter-settings ($\bm{x}$) to performance on the corresponding dataset. Our goal at each time step $t$ is then to find the best model-and-hyperparameter-setting, $\bm{x}_{t}^{*}$, that maximises $f_t$. More specifically, we look to find a maximiser of the $J$-fold cross-validation performance given by:
\begin{equation} \label{eq:obj}
\bm{x}_{t}^{*} \in \argmax_{\bm{x}_{t} \in \mathbb{X}} \frac{1}{J}\sum_{j=1}^{J} \mathcal{L}(\bm{x}_{t},\mathcal{D}_{t,train}^{(j)},\mathcal{D}_{t,valid}^{(j)})
\end{equation}
where $\mathcal{L}$ is some given performance metric (e.g. AUC-ROC, Model Likelihood, etc), and $\mathcal{D}_{t,train}^{(j)}$ and $\mathcal{D}_{t,valid}^{(j)}$ are training and validation splits of $\mathcal{D}_{t}$ in the $j$th fold, respectively. The objective in (\ref{eq:obj}) has no closed form solution (both the performance metric and the input space $\mathbb{X}$ are allowed to be anything), and so we must treat this as a black-box optimization problem. For each time step, we denote by $\mathcal{A}_t =(\mathbf{X}_t, \mathbf{Y}_t)= \{(\bm{x}_{i,t} , y_{i,t})\}_{i=1}^{N_t}$ the acquisition set for $f_t$ where $N_t$ is the number of acquisitions made for the $t$-th black-box function. We define the set of all acquisitions until time $t$ to be $\mathcal{A}_{\leq t} = \bigcup_{u=1}^{t}\mathcal{A}_u$. In a setting where datasets that are being collected are similar, our goal is to leverage past acquisitions, $\mathcal{A}_{\leq t-1}$, when learning $f_t$.

\begin{figure*}[h]
	\begin{center}
		\centerline{\includegraphics[width=12.70cm]{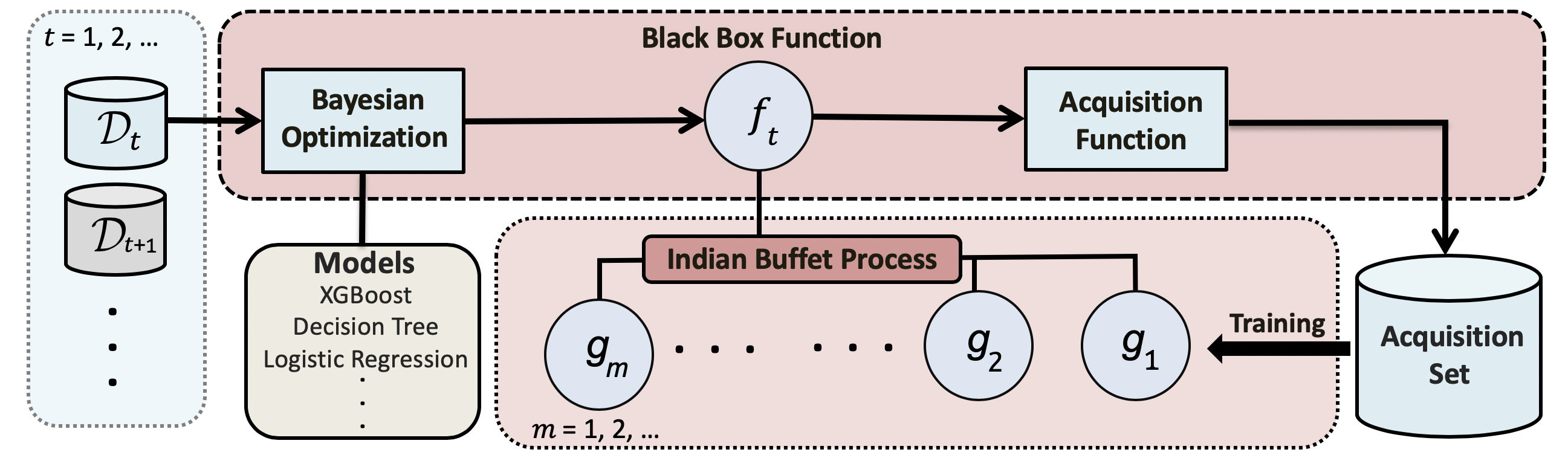}}
		\caption{A pictorial depiction of Lifelong Bayesian Optimization.  As the datasets arrive over time, the cross-validation performance is treated as a black-box function $f_t$. Some latent functions $g_m$ are generated in an Indian Buffet Process and trained on the acquisition set to fit $f_t$.}
		\label{nn1}
	\end{center}
\vskip -0.2in
\end{figure*}

In Linear Model of Coregionalization (LMC) \cite{Andre1978,Pierre1997,Yee05}, each black-box function $f_t$ is a linear combination of some latent functions, $g_1, ..., g_M : \mathbb{X} \to \mathbb{R}$ where each $g_m(\cdot)$ is drawn from a one-dimensional Gaussian process $\mathcal{GP}_m(0,K_m(\cdot,\cdot))$. The number of latent functions $M$ and the number of black-box functions $T$ are fixed. In \textsc{LBO}, $T$ and $M$ can be arbitrarily large and change over time. To allow for a suitable number of latent functions in our model, we set $M$ to be some large number and introduce an additional binary variable $z_{t,m}$ to determine whether the latent function $g_m(\cdot)$ is used in the modelling of $f_t(\cdot)$. $f_t$ is then defined by
\begin{equation*}
f_t(\cdot) = \sum_{m=1}^{M}z_{t,m} s_{t,m}g_m(\cdot)
\end{equation*}
where  $s_{t,m}$ is the weight (to be learned) of the latent function $g_m$ in the spanning of $f_t$. The correlation between $f_t$ and $f_{t'}$ is given by
$\mathrm{Cov}(f_t(\cdot),f_{t'}(\cdot)) = \sum_{m=1}^{M}z_{t,m}z_{t',m}s_{t,m}s_{t',m}K_m(\cdot,\cdot)$. This correlation is exploited to speed up the optimization process in Multitask BO \cite{SwerskySA13,KleinFBHH17}. The number of latent functions that have been used until a given time-step $t$ is given by $\tilde{M}_t = \sum_{m=1}^{M} \min(\sum_{u=1}^{t}z_{u,m},1)$. In \textsc{LBO}, as shown in Figure \ref{nn1}, the latent functions are generated by a Indian Buffet Process in the modelling of black-box functions. When a new black-box function $f_{t}$ can be spanned by the existing (i.e. previously used) latent functions, no new latent functions will be added to the model (i.e. $z_{t+1,m} = 0$ for $m > \tilde{M}_t$). We aim to learn the number of latent functions $\tilde{M}_t$, which allows us to identify the actual number of unknown functions in a sequence of black-box functions and avoid spending computational power in re-solving similar optimization problems that we encounter over time. Furthermore, when there is concept drift in the new arriving data set, we can model the corresponding black-box function independently by a new latent function, without enforcing it to be correlated with the previous black-box functions.

\subsection{Approximating latent processes using Neural Networks}\label{section:2}
To enable lifelong Bayesian optimization, the algorithm needs to be scalable to a large number of acquisitions over time. We consider the approach of approximating the infinite-dimensional feature map corresponding to the kernel of each latent Gaussian process $\mathcal{GP}_m(0,K_m(\cdot,\cdot))$, with a finite-dimensional feature map learned by Deep Neural Networks \cite{SnoekRSKSSPPA15,perrone2018scalable, springenberg2016bayesian}. Note that this approximation varies every time we optimize a new black-box function. When modelling $f_t$, we obtain a $D^{*}$-dimensional feature map $\bm{\phi}_{t,m}:\mathbb{X}\rightarrow\mathbb{R}^{D^{*}}$, where $D^{*}$ is the output dimension of the feature map. We set the dimension $D^{*}$ to be the same for all $\bm{\phi}_{t,m}, m=1, ..., M$. We then compose this with a final, fully connected layer $\mathbf{h}_{t,m} : D^* \to \mathbb{R}$ to define $g_{t,m}$ (i.e. $g_{t,m} = \mathbf{h}_{t,m}^{\top} \bm{\phi}_{t,m}$). We collect the parameters of each feature map, $\bm{\phi}_{t,1}, ..., \bm{\phi}_{t,m}$ into a single vector $\mathbf{\Theta}_{t}$. In our approximation $f_t(\cdot)\approx \sum_{m=1}^{M}z_{t,m} s_{t,m} \mathbf{h}_{t,m}^{\top} \bm{\phi}_{t,m}(\cdot)$, we can treat $s_{t,m}$ as part of $\mathbf{h}_{t,m}$ in learning and discard the parameters $s_{t,m}$ from our model. Let $\mathbf{\Phi}_t$ be the $N_t \times D^{*}M$ matrix s.t. $[\mathbf{\Phi}_t]_{u,m} =z_{u,m} \bm{\phi}_{t,m}(\bm{x}_u)$. We will also write $\bm{\phi}_{t,1:M}(\bm{x})$ to denote $[z_{t,1}\bm{\phi}_{t,1}^{\top}(\bm{x}),...,z_{t,M}\bm{\phi}_{t,M}^{\top}(\bm{x})]^{\top}$. Let $\mathbf{H}_t=\Big[\mathbf{h}_{t,1}^{\top},..., \mathbf{h}_{t,M}^{\top}\Big]^{\top}$, $\mathbf{Z}_t = [z_{t, 1}, ..., z_{t, M}]^{\top}$. Then the posterior distribution of $\mathbf{H}_t$ is given by
\begin{equation} \label{eq:h_post}
P(\mathbf{H}_t|\mathcal{A}_{t},\mathbf{\Theta}_t,\mathbf{Z}_t)= \mathcal{N}(\mathbf{m}_{\mathbf{H}_t},\mathbf{K}_{\mathbf{H}_t}^{-1})
\end{equation}
where the mean function, $\mathbf{m}_{\mathbf{H}_t}$, and the kernel, $\mathbf{K}_{\mathbf{H}_t}$, are given by 
\begin{equation*}
\mathbf{m}_{\mathbf{H}_t}=\frac{\beta_t}{\lambda_t} \mathbf{K}_{\mathbf{H}}^{-1}\mathbf{\Phi}_t^{\top}\mathbf{Y}_t,
\ \ \ \mathbf{K}_{\mathbf{H}_t} =\frac{\beta_t}{\lambda_t} \mathbf{\Phi}_t^{\top}\mathbf{\Phi}_t+\mathbf{I}_{D^{*}\times D^{*}}
\end{equation*}
where $\lambda_t$ is the precision parameter in the prior distributions $P(\mathbf{h}_{t,m}|\lambda_t)=\mathcal{N}(\mathbf{0},\lambda_{t}^{-1}\mathbf{I}_{D^{*}\times D^{*}})$, and $\beta_t$ is the precision parameter in the likelihood function
\begin{equation}\label{equ:likelihood}
    \begin{split}
    P(\mathbf{Y}_t|\mathbf{f}_t,\beta_t) = P(\mathbf{Y}_t|\mathbf{Z}_t,\mathbf{X}_t,\mathbf{H}_t, \mathbf{\Theta}_t )  =\prod_{i=1}^{N_t}\mathcal{N}(y_{i,t};\sum_{m=1}^{M}z_{t,m}\mathbf{h}_{t,m}^{\top} \bm{\phi}_{t,m}(\bm{x}_{i,t}),\beta_{t}^{-1})
    \end{split}
\end{equation}
In \cite{SnoekRSKSSPPA15}, the neural network parameters  $\mathbf{\Theta}_t,\mathbf{H}_t$ are optimized end-to-end by stochastic gradient descent. This approach is shown to be disadvantageous in the setting of Multitask BO \cite{perrone2018scalable}. In \textsc{LBO}, we only learn the parameter $\mathbf{\Theta}_t$ by marginal likelihood optimization. The parameter $\mathbf{H}_t$ is integrated out w.r.t its posterior distribution from (\ref{eq:h_post}) when we derive the predictive distribution for a testing data point $\bm{x}$ as
\begin{equation*}
f_t(\bm{x}) \sim \mathcal{N}\big(\mu(\bm{x}|\mathcal{A}_{t},\mathbf{\Theta}_t,\mathbf{Z}_t),\sigma^2(\bm{x}|\mathcal{A}_{t},\mathbf{\Theta}_t,\mathbf{Z}_t)\big)
\end{equation*}
where 
\begin{equation}\label{equ:mean_var}
\mu(\bm{x}|\mathcal{A}_{t},\mathbf{\Theta}_t,\mathbf{Z}_t) = \mathbf{m}_{\mathbf{H}_t}^{\top}\bm{\phi}_{t,1:M}(\bm{x}), \ \ \  \sigma^2(\bm{x}|\mathcal{A}_{t},\mathbf{\Theta}_t,\mathbf{Z}_t)  = \bm{\phi}_{t,1:M}(\bm{x})^{\top}\mathbf{K}_{\mathbf{H}_t}^{-1}\bm{\phi}_{t,1:M}(\bm{x})
\end{equation}
\textbf{Complexity.} In marginal likelihood optimization, we switch between the primal and dual form of the log marginal likelihood for computational efficiency and numerical stability. In primal form, the log marginal likelihood is
\begin{equation*}\label{equ:marginal_primal}
    \begin{split}
    \mathcal{L}( \mathbf{\Theta}_t )=-\frac{N_t}{2}\log (2\pi\beta_t^{-1}) - \frac{\beta_t}{2}\|\mathbf{Y}_t\|^{2} + \frac{\beta_t^2}{\lambda_t}\mathbf{Y}_t^{\top}\mathbf{\Phi}_t \mathbf{K}_{\mathbf{H}_t}^{-1} \mathbf{\Phi}_t^{\top}\mathbf{Y}_t - \frac{1}{2}\log|\mathbf{K}_{\mathbf{H}_t}|
    \end{split}
\end{equation*}
In dual form, it is given as the logarithm of $\mathcal{N}(\mathbf{Y}_t;\lambda_t^{-1}\mathbf{\Phi}_t\mathbf{\Phi}_t^{\top}+\beta_t^{-1}\mathbf{I}_{N_t\times N_t})$. When $N_t>\tilde{M}_tD^{*}$, we optimize the log marginal likelihood in primal form, otherwise in dual form. Recall that $\tilde{M}_t$ denotes the number of active neural networks. The dimension of the feature map $\bm{\phi}_{t,1:M}(\bm{x})$ is equal to $\tilde{M}_tD^{*}$ since some $z_{t,m}$ are zero. Despite the fact that we set a large value for $M$, the complexity only depends on the lesser of the number of active neural networks and the size of the acquisition data. Overall, optimizing the log marginal likelihood has complexity $O(\max(\tilde{M}_t D^{*},N_t)(\min(\tilde{M}_t D^{*},N_t))^2)$. When we initialize $\textsc{LBO}$ by activating a relatively large number of neural networks, the computational overhead is still affordable since the acquisition data is small. When we acquire more and more data, we use an Indian Buffet Process to keep a small number of activated neural networks.

\subsection{Modelling $\mathbf{Z}$ as an Indian Buffet Process}
We put an Indian Buffet Process (IBP) prior on the binary matrix $\mathbf{Z}$, where $[\mathbf{Z}]_{t,m}=z_{t,m}$ and $z_{t,m} = 1$ if the $m$th neural network $g_{m}(\cdot)$ is used to span the $t$th black box function and $z_{t,m} = 0$ otherwise. In the Stick-breaking Construction of IBP \cite{TehGG07}, a probability $\pi_m\in[0,1]$ is first assigned to each column of $\mathbf{Z}$. In the $t$th column $\mathbf{Z}_t$, each $z_{t,m}$ is sampled independently from the distribution $\Bernoulli(\pi_m)$. The sampling order of the rows does not change the distribution. The probability $\pi_m$ is generated according to
\begin{equation} \label{equ:IBP}
\pi_m =  \prod_{k=1}^{m}v_k \text{ where } v_k \sim \Beta(\alpha,1)
\end{equation}
In this construction, $\mathbb{E}[\pi_m]$, the expected probability of using the $m$th neural network, decreases exponentially as $m$ increases. The parameter $\alpha$ represents the expected number of active neural networks for each black-box function. It controls how quickly the probabilities $\pi_m$ decay. Larger values of $\alpha$ correspond to a slower decrease in $\pi_m$, and thus we would expect to use more neural networks to represent the black box functions. Using IBP we are able to limit the number of neural networks used at each time-step, while also introducing new neural networks when the new black-box function is distinct to the previous ones. 

\subsubsection{Mean Field Approximation} \label{sec:varinf}
We use the variational method \cite{singh2017structured,nalisnick2016stick} to infer $\mathbf{Z}_t$, which involves a mean-field approximation to the posterior distribution of $\mathbf{Z}_t$ and learning the variational parameters through the evidence lower bound (ELBO) maximization. The joint distribution of the acquisitions $(\mathbf{X}_t,\mathbf{Y}_t)$ and latent variables $\mathbf{Z}_t$ is given by
\begin{equation*}
P(\mathbf{Y}_t,\mathbf{Z}_t,|\mathbf{X}_t,\mathbf{\Theta}_t,\mathbf{v})  = P(\mathbf{Y}_t|\mathbf{Z}_t,\mathbf{X}_t,\mathbf{\Theta}_t)P(\mathbf{Z}_t|\mathbf{v})
\end{equation*}
where $\mathbf{v} = \Big[\pi_{1},...,\pi_{M}\Big]$. The prior distribution of $\mathbf{Z}_t$ is given by $P(\mathbf{Z}_t|\mathbf{v})= \prod_{m=1}^{M}\Bernoulli(\pi_m)$. Since computing the true posterior distribution $P(\mathbf{Z}_t|\mathbf{X}_t,\mathbf{Y}_t,\mathbf{\Theta}_t)$ is intractable, we use a fully factorized mean-field approximation given by $Q(\mathbf{Z}_t) = \prod_{m=1}^{M}Q_{\gamma_{t,m},\tau_{t,m}}(z_{t,m})$ where $Q_{\gamma_{t,m},\tau_{t,m}}(z_{t,m})$ is a Binary Concrete distribution, the continuous relaxation of the discrete Bernoulli distribution,  
\begin{equation}\label{BinConcrete}
  \BinConcrete(\gamma,\tau) = \frac{\gamma\tau z^{-\tau-1}(1-z)^{-\tau-1}}{(\gamma z^{-\tau}+(1-z)^{-\tau})^2}
\end{equation}
where $z\in(0,1)$, $\gamma\in(0,\infty)$ is a probability ratio and $\tau\in(0,\infty)$ is the temperature hyperparameter. As $\tau\rightarrow 0$, samples from the Concrete distribution are binary and identical to the samples from a discrete Bernoulli distribution. A realization of the Concrete variable $z$ has a convenient and differentiable parametrization:
\begin{equation}\label{equ:concrete}
    z = \Sigmoid \Big(\frac{1}{\tau} \big(\log\gamma-\log(u)-\log(1-u)\big)\Big)
\end{equation}
where $u\sim\Uniform(0,1)$. We can learn $\mathbf{\Theta}_t$ and  $\{\gamma_{t,m}\}_{m=1}^{M}$ jointly by maximizing the ELBO of the log-marginal likelihood given by 
\begin{equation}\label{equ:ELBO}
\mathcal{L}_{e}(Q,\mathbf{\Theta}_t) =\big\langle\log P(\mathbf{Y}_t|\mathbf{X}_t,\mathbf{Z}_t,\mathbf{\Theta}_t) \big\rangle_{Q(\mathbf{Z}_t)} - D_{KL}\big[Q(\mathbf{Z}_t)||P(\mathbf{Z}_t|\mathbf{v})\big]
\end{equation}
where $D_{KL}[\cdot\|\cdot]$ is the Kullback–Leibler divergence. We approximate the expectation and KL divergence in (\ref{equ:ELBO}) by drawing a sample $z_{t,m}$ in (\ref{equ:concrete}) and $\pi_m$ in (\ref{equ:IBP}). Since $z_{t,m}$ is an approximate binary sample, we cannot evaluate it in the Binary distribution $\Bernoulli(\pi_m)$. Therefore, we also approximate each $\Bernoulli(\pi_m)$ in $P(\mathbf{Z}_t|\mathbf{v})$ by the corresponding Concrete distribution $\BinConcrete(\frac{\pi_m}{1-\pi_m},\tau)$ when evaluating the KL divergence in (\ref{equ:ELBO}). Intuitively, ELBO maximization has two objectives: (1) Tuning the feature map parameter $\mathbf{\Theta}_t$ to model the data $(\mathbf{X}_t,\mathbf{Y}_t)$; and (2) Learning $Q(\mathbf{Z}_t)$ to infer how many neural networks are required to achieve the first objective with a small KL divergence from $P(\mathbf{Z}_t|\mathbf{v})$.

\subsection{Online Training of Neural networks} \label{sec:onltra}

In the update of $\mathbf{\Theta}_t$, we use only the data (acquisitions) from the current black-box function $f_t$. To prevent catastrophic forgetting and leverage the past data, we can use the neural network weights from previously learned neural networks as a graph regularizer \cite{micchelli2005kernels,sheldon2008graphical,kirkpatrick2017overcoming,zenke2017continual}. In \textsc{LBO}, we use the unweighted graph regularizer in \cite{sheldon2008graphical}. Let $\mathbf{W}_{u,m}$ denote the weight parameters in $\bm{\phi}_{u,m}(\cdot)$ when we finish optimizing the $u$th black-box function. Then, at time $t$, we optimize the ELBO in (\ref{equ:ELBO}) with the graph regularizer on $\mathbf{W}_{t,1:M}$ being given by
\begin{equation} \label{eq:reg}
\begin{split}
\mathbf{\Omega}(\mathbf{W}_{t,1:M};\mathbf{W}_{1:t-1,1:M},\mathbf{Z}) = 
 \sum_{m=1}^{M}\Big(\sum_{u=1}^{t-1}z_{t,m}z_{u,m} \|\mathbf{W}_{t,m} - \mathbf{W}_{u,m}\|_{2}^2 \Big)
\end{split}
\end{equation}
The regularizer enforces that the current weights $\mathbf{W}_{t,1:M}$ for each neural network are similar to the previous weights $\mathbf{W}_{1:t-1,1:M}$ with the term $z_{u,m} z_{t,m}$ ensuring that similarity is only enforced between two time-steps that actually use the neural networks $m$ in their decomposition of $f$. Our primary interest is $f_t$ rather than the previous black-box functions that have already been optimized and so we allow each neural network to deviate slightly from the optimal solution for previous black-box functions. If the ELBO can be optimized under this constraint, $f_t$ is said to neighbour some of the previous black-box functions. If the existing neural networks cannot fit the new black-box function well, then when updating $\mathbf{Z}_{t}$, and a new untrained neural network will be used in addition to the existing ones. This new network will not have any similarity constraints since its corresponding penalty term in (\ref{eq:reg}) will be zero. Pseudo-code for the algorithm can be found in the Supplementary Materials.

\subsection{Hyperparameters}
In Section 2.1, the posterior distribution $P(\mathbf{H}_t|\mathcal{A}_t,\mathbf{\Theta}_t,\mathbf{Z}_t)$ has two precision parameters $\lambda_t$ and $\beta_t$. We learn these two hyperparameters by ELBO maximization i.e. Empirical Bayes \cite{williams2006gaussian}. In Section 2.2.1, the Binary Concrete distribution has a temperature parameter of $\tau$. As shown in the works \cite{maddison2016concrete,jang2016categorical}, Setting $\tau$ to $0.1$ is sufficient to obtain approximate one-hot samples from a discrete Bernoulli distribution. In the Indian Buffet Process, $\alpha$ is the expected number of active neural networks. Recall that  $D_{*}$ is the input dimension of each feature map $\bm{\phi}_{t,m}$ defined in Section \ref{section:2}. The parameter $\alpha$ is set to be a small positive number such that it is computationally affordable to invert $\alpha D_{*} \times \alpha D_{*}$ matrices when optimizing the ELBO in (\ref{equ:ELBO}). In the following section of experiments, we set $\alpha$ to be 2 when the feature maps are fifty-dimensional. Following the work \cite{sheldon2008graphical}, we select the regularization parameter for (\ref{eq:reg}) by a grid search over $\{{1,10^{-1},10^{-2},10^{-3}},10^{-4}\}$, choosing the parameter with the best cross-validation error on the acquisition data.

\section{Experiments}
In this section, we assess the ability of \textsc{LBO} to optimize a sequence of black-box functions. We first examine the effectiveness and robustness of the algorithm on a simple synthetic data set. We compare \textsc{LBO} with Random Search, Multitask GP (\textsc{MTGP}) \cite{SwerskySA13}, Multitask Neural Network model \textsc{ABLR} \cite{perrone2018scalable}, Single-task GP model (\textsc{SGP}) \cite{snoek2012practical} and Single-task Neural Network model \textsc{DNGO} \cite{SnoekRSKSSPPA15}. We then compare the BO algorithms on three publicly available UNOS datasets\footnote{https://unos.org/data/}.

\subsection{Synthetic dataset}\label{section:Synthetic}

In the synthetic dataset, we consider five sequences of five Branin functions that need to be minimized. The standard Branin function is $f(\bm{x}) = a(x_2-b x_1^{2}+c x_1-r)^2+s(1-r)\cos(x_1)+s$ where $a=1,b=5.1/(4\pi^2),c=5/\pi,r=6,s=10$ and $t=1/(8\pi)$. This function is evaluated on the square $x_1 \in [-5, 10], x_2 \in [0, 15]$. Each sequence consists of five Branin functions $f_{t,1},...,f_{t,5}$. We create each $f_{t,i}$ by adding a small constant to each parameter of the standard Branin function. The constant is drawn from $\mathcal{N}(0, \sigma_{t}^2)$, where $\sigma_{1}=0.01,\sigma_{2}=0.05,\sigma_{3}=0.1,\sigma_{4}=0.5$ and $\sigma_{5}=1$. The correlation between functions decreases with increasing $\sigma_t$. Each of the sequences are optimized independently. We aim to test the robustness of different BO algorithms when they are applied to optimize sequences of functions at different levels of correlation. In \textsc{LBO}, we set the maximum number of neural networks, $M$, to 10. Each neural network $g_1,..., g_5$ has three layers with fifty $\tanh$ units. \textsc{LBO} is optimized using Adam \cite{kingma2014adam}. The benchmark \textsc{DNGO} is a single neural network, with three layers each with fifty $\tanh$ units. \textsc{ABLR} is a Multitask neural network, also with three layers with $\tanh$ activations. We make sure the neural network is sufficiently wide that it can learn the shared feature map for the black-box functions. We conduct the experiments for \textsc{ABLR} with fifty and one-hundred units, denoted by \textsc{ABLR-50} and \textsc{ABLR-100} respectively. The same architectures are used in the original paper \cite{perrone2018scalable} after investigating the computation time of ABLR. The benchmark \textsc{SGP} is a Gaussian process implemented using the \texttt{GPyOpt} library \cite{gonzalez2016gpyopt}. The benchmark \textsc{MTGP} is implemented with a full-rank task covariance matrix, using the \texttt{GPy} library \cite{gpy2014}. Both GP models use a Matérn-5/2 covariance kernel and automatic relevance determination hyperparameters, optimized by Empirical Bayes \cite{williams2006gaussian}. For every algorithm, we use Expected Improvement (EI) \cite{jones1998efficient} as the acquisition function $a_f$. We set the maximum number of function evaluations to 200. All the BO optimizers are initialized with the same five random samples in the domain of the Branin function.

\begin{figure}[ht]
	\vskip -0.1in
  \subfloat[]{
	\begin{minipage}{
	   0.32\textwidth}
	   \includegraphics[height=3.8cm, width=0.95\linewidth]{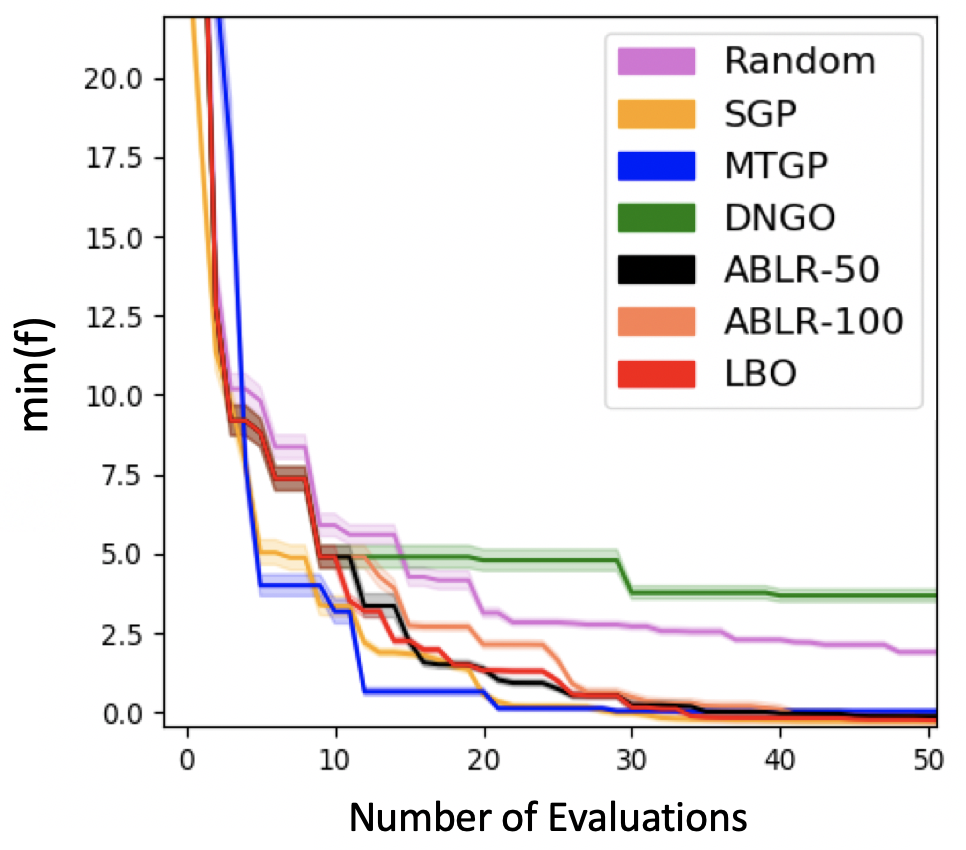}
	   \label{fig:seq123}
	\end{minipage}}
  \subfloat[]{
	\begin{minipage}{
	   0.32\textwidth}
	   \includegraphics[height=3.8cm, width=0.95\linewidth]{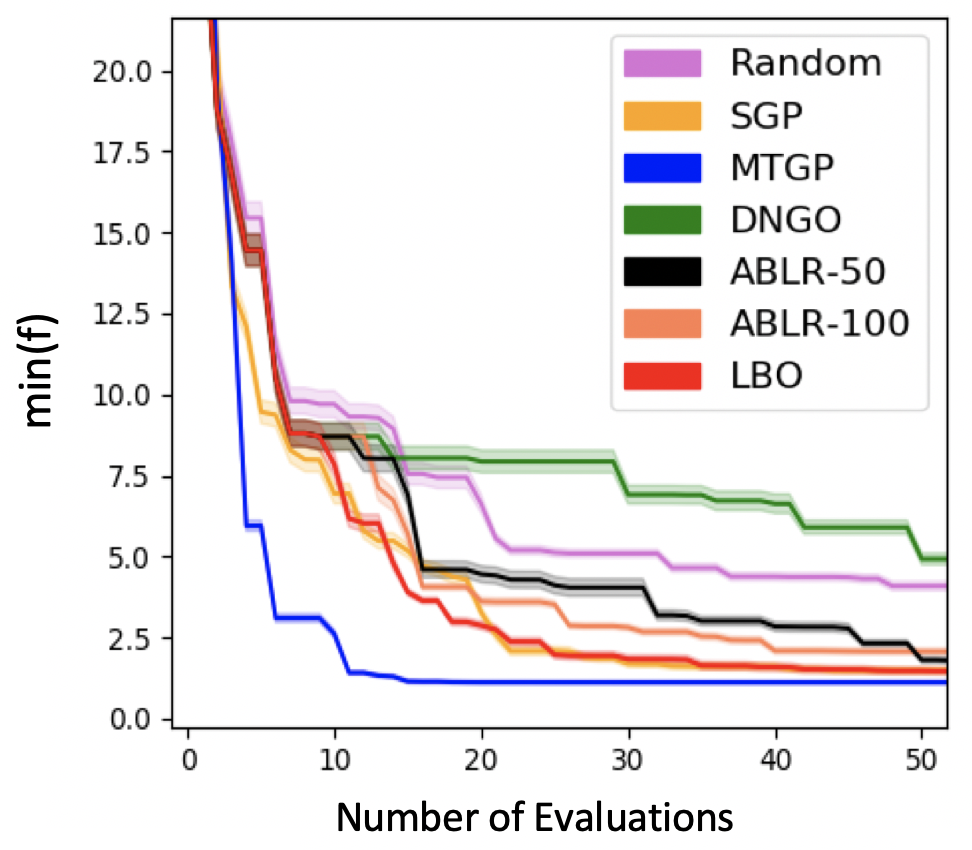}
	   \label{fig:seq345}
	\end{minipage}}
	  \subfloat[]{
	\begin{minipage}{
	   0.32\textwidth}
	   \includegraphics[height=3.8cm, width=0.97\linewidth]{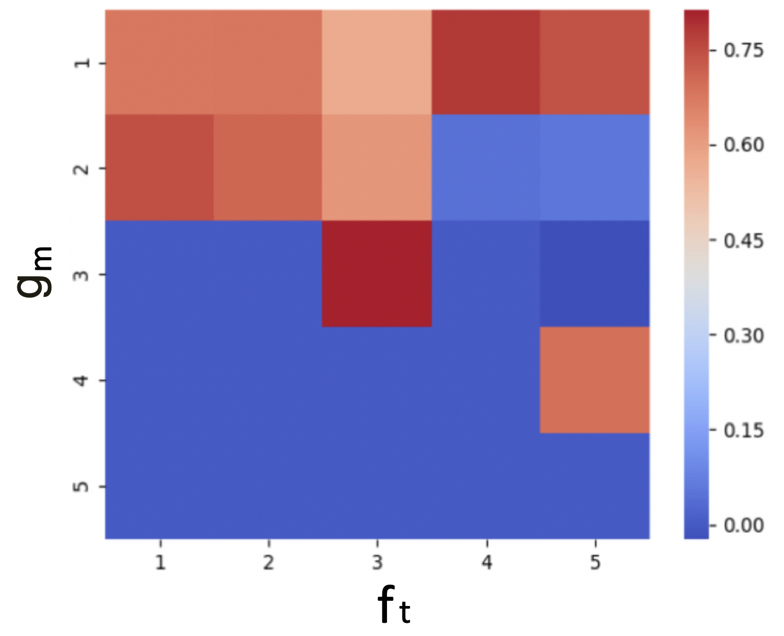}
	   \label{fig:seq_corr}
	\end{minipage}}
\caption{(a) The average minimum function values in the first fifty evaluations for the sequence $t=  1, 2, 3$; (b) The average minimum function values in the first fifty evaluations for the sequence $t = 3, 4, 5$; (c) Correlation between neural networks $g_1, ..., g_5$ and black-box functions $f_1, ..., f_5$.}
\label{fig:nonlinear}
	\vskip -0.1in
\end{figure}

The experiments of each sequence are averaged over 10 random repetitions. Figure \ref{fig:seq123} shows the minimum function values averaged over the sequence $t=1,2,3$. It represents the performance of the BO algorithms when the function correlation is strong. Figure \ref{fig:seq345} shows the same results for the sequence $t=3,4,5$, which compares the BO algorithms when the function correlation is weak. As expected, for a smooth function like Branin, \textsc{MTGP} and \textsc{SGP} take fewer samples to approximate it and find its minimum because GP models start with the right kernel to model the smooth function while neural networks take more samples to learn the kernel from scratch. In \textsc{ABLR}, all the black-box function share one single neural network. The size of the shared network and the function correlation determine the sample complexity of neural network based BO. Comparing the results of \textsc{ABLR-50} and \textsc{ABLR-100}, we find the former converges faster than the latter when function correlation is strong (as in Figure \ref{fig:seq123}), while the converse is true when function correlation is weak (as in Figure \ref{fig:seq345}). Selecting the size of the neural networks requires prior knowledge of the correlation between the black-box functions. In Figure \ref{fig:seq123}, the performance of \textsc{LBO} is comparable to \textsc{ABLR-50} when the correlated functions share the same feature map. In Figure \ref{fig:seq345}, \textsc{LBO} outperforms \textsc{ABLR-50} and \textsc{ABLR-100}, by learning the latent variable $z_{t,m}$ to infer a suitable number of neural networks in modelling the black-box functions, and which functions should share the same feature maps. To further understand the results of \textsc{LBO}, we compute the correlation matrix between the neural networks $g_m$ and the noisy observation $y_t$ of $f_t$. The sample correlation is estimated as
\begin{equation} \label{eq:heatmap}
\hat{\Corr}= \frac{\sum_{j=1}^{N_t}(g_m(\bm{x}_{j,t})-\bar{g}_m)(y_{j,t}-\bar{y}_{t}) }{ \sqrt{\sum_{j=1}^{N_t}\big(g_m(\bm{x}_{j,t})-\bar{g}_m\big)^2
		\sum_{j=1}^{N_t}\big(y_{j,t}-\bar{y}_{t}\big)^2 }}
\end{equation}
where $\bar{y}_{t}$ and $\bar{g}_m$ are the sample means. In Figure \ref{fig:seq_corr}, we show the correlation multiplied by $z_{t,m}$. We see from Figure \ref{fig:seq_corr}, that $g_{1}$ is used by all the black-box functions. We also see that both of $f_1$ and $f_2$ can be fit using the same two neural networks, $g_1$ and $g_2$ but the third function, $f_3$ requires a new neural network, $g_3$ to be introduced in order to find a good fit. This indicates that there were no neighbours of $g_1$ and $g_2$ that would suitably fit $f_3$.

\subsection{Real-world datasets}\label{section:Real}
We extracted the UNOS-I, UNOS-II and UNOS-III data between the years 1985 to 2004 from the United Network for Organ Sharing (UNOS). Cohort UNOS-I is a pre-transplant population of cardiac patients who were enrolled in a heart transplant wait-list. Cohort UNOS-II is a post-transplant population of patients who underwent a heart transplant. Cohort UNOS-III is a post-transplant population of patients who underwent a lung transplant. We divide each of the datasets into subsets $\mathcal{D}_{t},t=1,2,...,T$, varying the number of $T$ throughout the experiments. The model is tasked with predicting survival at the time horizon of three years. As each $\mathcal{D}_{t}$ arrives over time, we treat the 5-fold cross-validation AUC of the models on $\mathcal{D}_{t}$ as a black-box function $f_t$.  In our experiment, the models we consider for selection are XGBoost \cite{chen2016xgboost}, Logistic Regression \cite{mccullagh1989generalized}, Bernoulli naive bayes \cite{mccallum1998comparison}, Multinomial naive bayes \cite{kibriya2004multinomial}. XGBoost is implemented with the Python package \texttt{xgboost}. Other models are implemented with the \texttt{scikit-learn} library \cite{scikit-learn}. The input space $\mathbb{X}$ is the product of all hyperparameter spaces for all the models together with a categorical variable indicating which model is selected. The details of the full hyperparameter space are provided in the Supplementary Material.

\begin{figure}[h]
    \vskip -0.1in
  \subfloat[]{
	\begin{minipage}{0.31\textwidth}
	   \includegraphics[height=3.8cm, width=0.92\linewidth]{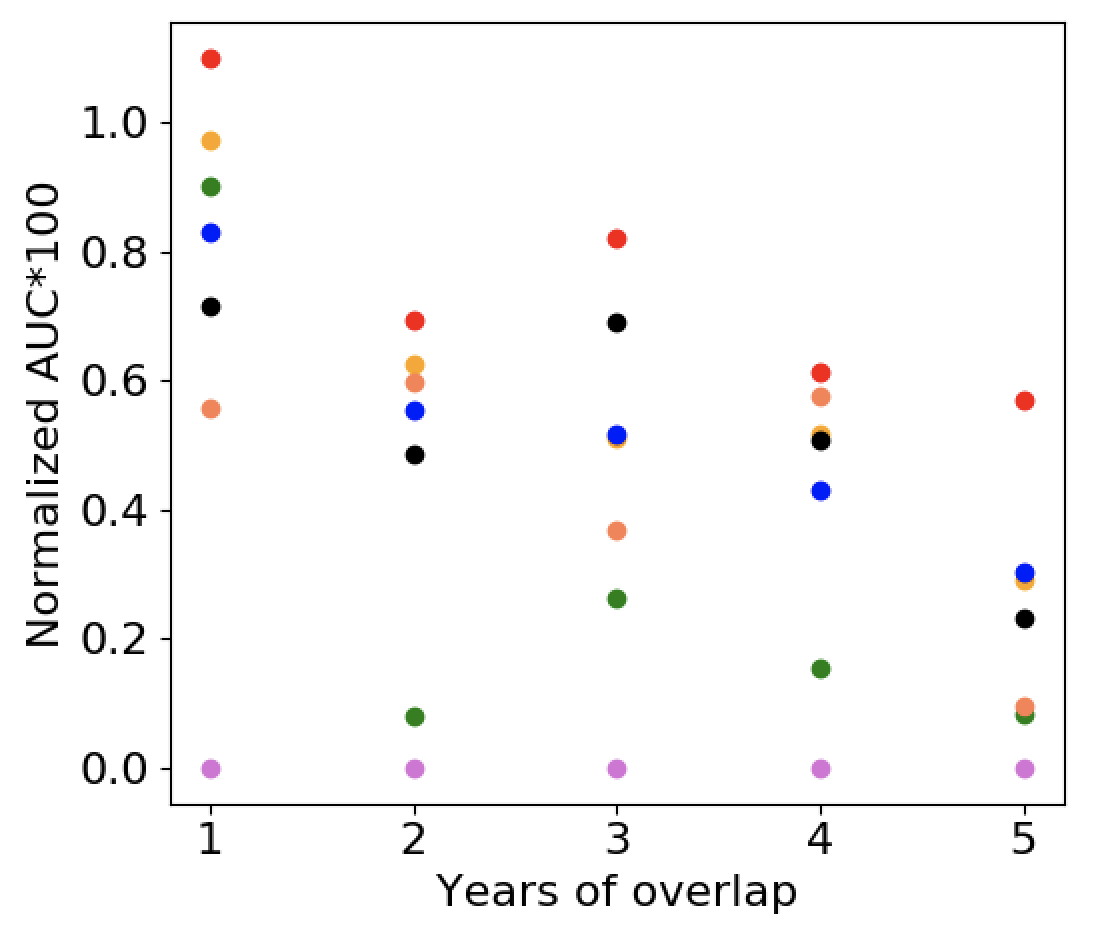}
	   \label{fig:UNOS1}
	\end{minipage}}
  \subfloat[]{
	\begin{minipage}{0.31\textwidth}
	   \includegraphics[height=3.84cm, width=0.94\linewidth]{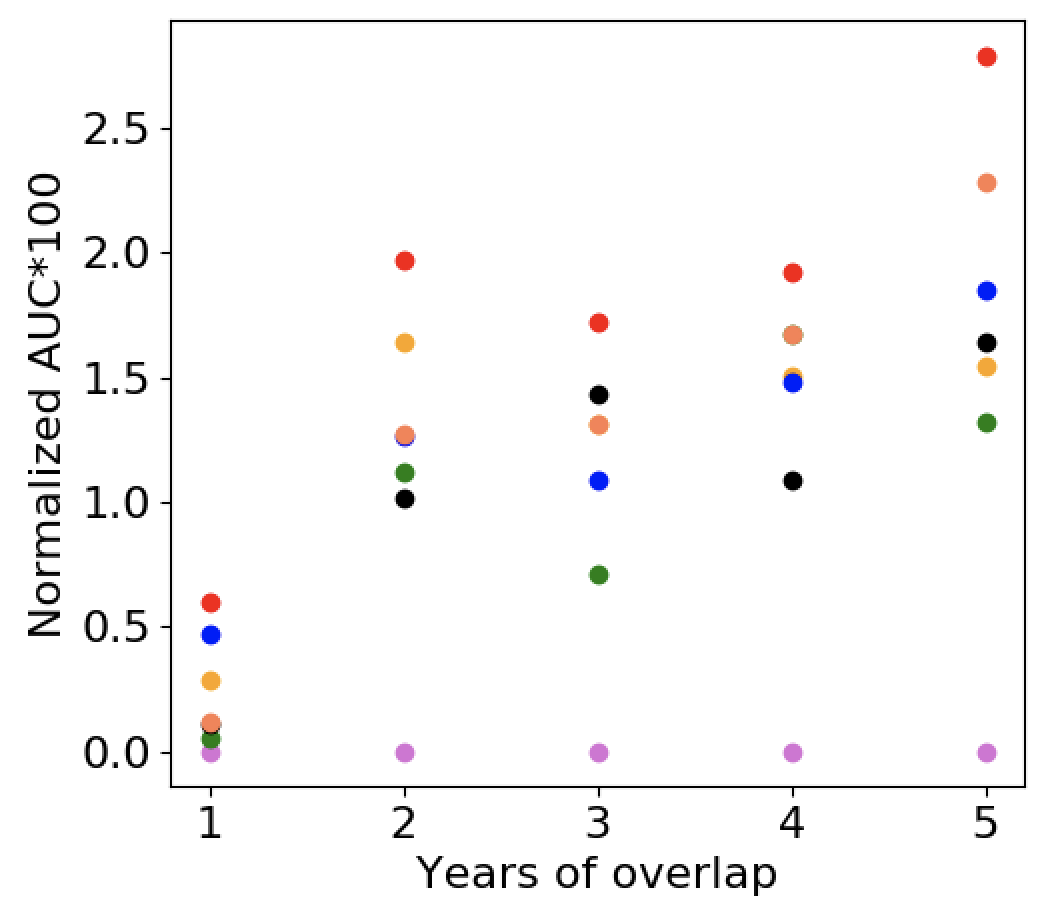}
	   \label{fig:UNOS2}
	\end{minipage}}
	  \subfloat[]{
	\begin{minipage}{0.38\textwidth}
	   \includegraphics[height=3.8cm, width=5cm]{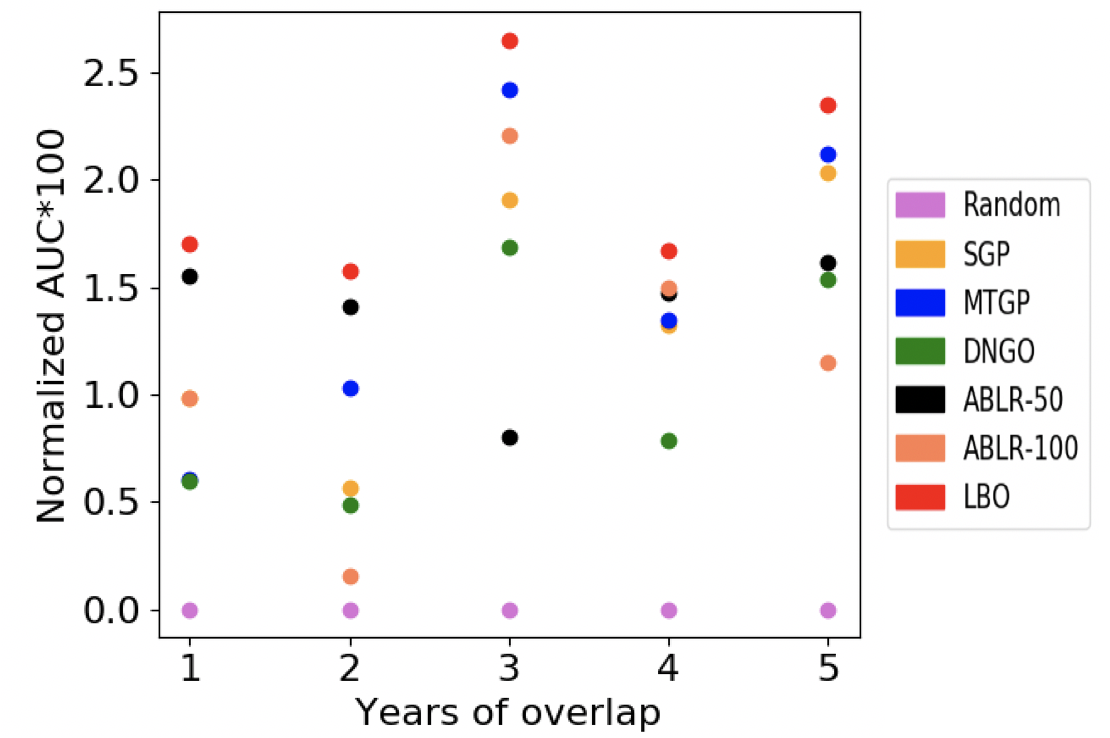}
	   \label{fig:UNOS2}
	\end{minipage}}
\caption{Normalized AUC after fifty function evaluations: (a) UNOS-I; (b) UNOS-II; (c) UNOS-III.}
\vskip -0.1in
\label{fig:AUC}
\end{figure}

To compare lifelong learning with multitask learning, we create varying amounts of \quotes{overlap} between each dataset. To do this, we set a year \quotes{width} (from 4, 5, 6) and then set the first dataset to be all data from 1985 until 1985+width. Each subsequent dataset is created by removing the oldest year from the previous dataset and adding in the next year (for example, the second dataset consists of all data from 1986 until 1986+width). The total number of years of overlap between each subsequent dataset is therefore width minus one. Note that this set-up corresponds to several standard data collection mechanisms in which data is collected regularly and \quotes{old data} is phased out (because it is believed to no longer be indicative of the current distribution, for example).

We compare \textsc{LBO} against Random Search, \textsc{SGP}, \textsc{MTGP}, \textsc{DGNO}, \textsc{ABLR-50} and \textsc{ABLR-100}. The models \textsc{LBO}, \textsc{DGNO}, \textsc{ABLR-50} and \textsc{ABLR-100} have the same architecture as in the Section \ref{section:Synthetic}. We set the maximum number of function evaluation to 200. All the BO optimizers are initialized with the same five random samples. Since the point of LBO is to increase the speed at which we learn the best model, the primary metric we report is maximum AUC after the first fifty function evaluations. In order to visualize and compare the results on the datasets with different overlap, we normalized the AUC of each BO algorithm by the AUC of Random Search (i.e. the AUC is first subtracted, then divided by the AUC of Random Search). Figure \ref{fig:AUC} shows the experimental results averaged over 10 random repetitions. In the experiments with different amounts of \quotes{overlap}, \textsc{LBO} is the most consistent method over all the experiments. \textsc{LBO} outperforms the benchmarks in the UNOS-I, II and III datasets for all overlaps. \textsc{LBO} shows the gain from learning a suitable number of neural networks to model the black-box functions, rather than enforcing them to share one single neural network as in \textsc{ABLR} or assuming the number of latent functions is equal to the number of black-box functions as in \textsc{MTGP}. Our method \textsc{LBO} offers extra robustness when leveraging past data to speed up the current optimization. In the Supplementary Materials, we provide the details of overlapping over the years and the results of 200 function evaluations for all datasets and overlaps.

\section{Conclusion}
This paper developed Lifelong Bayesian Optimization (\textsc{LBO}) to solve the model selection problem in the setting where datasets are arriving over time. \textsc{LBO} improves over standard BO algorithms by capturing correlations between black-box functions corresponding to different datasets. We used an Indian Buffet Process to control the introduction of new neural networks to the model. We applied a variational method to optimize our model such that the neural network parameters and the latent variables can be learned jointly. Through synthetic and real-world experiments we demonstrated that LBO can improve the speed and robustness of BO.
\small

\bibliographystyle{unsrt}
\bibliography{neurips_2019_ver40} 

\clearpage
\newpage

\section*{Supplementary Material}
\subsection*{Overlap Datasets}

\begin{figure}[h]
	\centering
	  \subfloat[]{
	\begin{minipage}{0.5\textwidth}
	   \centering
	   \includegraphics[height=3.5cm, width=8cm]{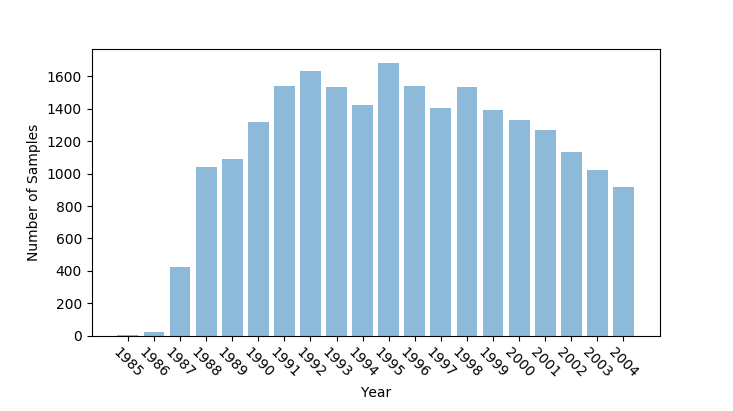}
	\end{minipage}}
	\hfill
	   \subfloat[]{
	\begin{minipage}{0.5\textwidth}
	   \centering
	   \includegraphics[height=3.5cm, width=8cm]{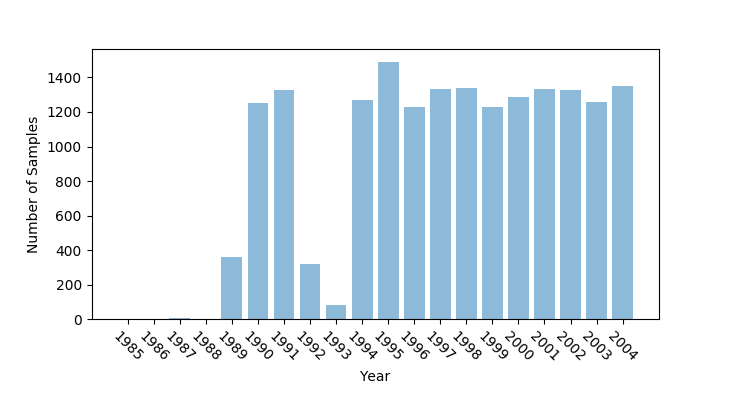}
	\end{minipage}}
	\hfill
	   \subfloat[]{
	\begin{minipage}{0.5\textwidth}
	   \centering
	   \includegraphics[height=3.5cm, width=8cm]{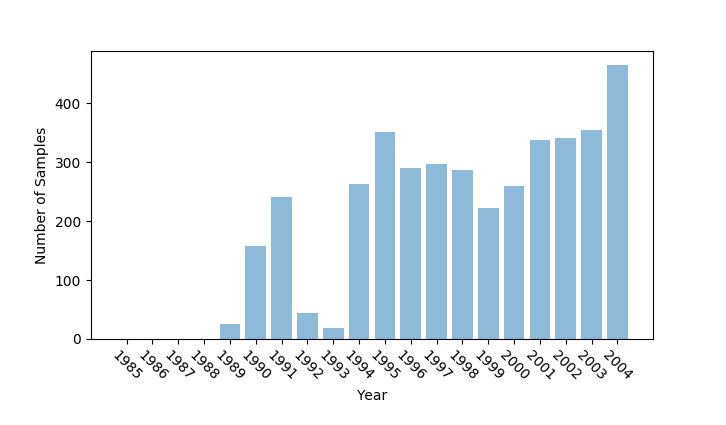}
	\end{minipage}}
\caption{Histograms of sample sizes in years: (a) UNOS-I; (b) UNOS-II; (c) UNOS-III.}
\label{fig:hist_UNOS}
\vskip -0.1in
\end{figure}

In Section \ref{section:Real}, we subsampled the UNOS datasets to create sequences of datasets with different amount of overlaps. In this section, we provide the details of the \quotes{overlaps} experiments, including histograms of sample sizes and tables of \quotes{overlaps} for each UNOS dataset. In all the UNOS datasets, we remove the samples with missing features. In Figure \ref{fig:hist_UNOS}, we show the histograms of sample sizes from 1985 to 2004. In UNOS-I, the samples are distributed between 1985 and 2004. In UNOS-II and III, the majority of samples are distributed between 1989 and 2004. Therefore, the \quotes{overlaps} experiments of UNOS-I start from 1985 while the experiments of UNOS-II and III begin with 1989. The details of \quotes{overlaps} are provided in Tables \ref{tab:UNOS-I}, \ref{tab:UNOS-II} and  \ref{tab:UNOS-III}. 

In each overlap experiment (i.e. each row of the tables), we move the windows five times, which creates a sequence of five overlapped datasets. For example, in the row \quotes{1 year} of Table \ref{tab:UNOS-I}, we create a sequence of five datasets with one-year overlap. We create the first dataset in this sequence by sub-sampling the UNOS-I data in the years between 1985 and 1988.  The complexity of MTGP \cite{SwerskySA13} and \textsc{ABLR} \cite{perrone2018scalable} increases with an increasing number of black-box functions in the sequence. Especially, we implemented \textsc{MTGP} without using a low-rank approximation of the task covariance matrix. It is expensive to extend our experiments to optimize a long sequence of functions since we need to retrain the Multitask BO models with all the previous acquisitions data every time we optimize a new black-box function. In this paper, we focus on comparing the BO algorithms in terms of their speed in finding the function maximizers or minimizers. Not only can \textsc{LBO} speed up the optimization process of each black-box function, but it is also scalable with an increasing number of black-box functions and their acquisition data. This is an important advantage of \textsc{LBO} when we use it to keep updating machine learning models or pipelines over time with new arriving data. With \textsc{LBO}, we can always leverage the past acquisition data in the optimization process of the current black-box function, despite the ever increasing size of the acquisition data we accumulated over time.

{
	\renewcommand{\arraystretch}{1.5}
	\begin{table*}[h]
		\centering
		\begin{small}
			\begin{tabular}{|c|c|c|c|c|c|}
				\hline
				 \backslashbox{Overlaps}{Years} 
                    & \multicolumn{5}{|c|}{UNOS-I}\\
				\hline
				1 year  & 1985-88 & 1988-91 & 1991-94 & 1994-97  & 1997-00 \\
				\hline
				2 years & 1985-88 & 1987-90 & 1989-92 & 1991-94  & 1993-96  \\
				\hline
				3 years & 1985-88 & 1986-89 & 1987-90 & 1988-91  & 1989-92 \\
				\hline
				4 years & 1985-89 & 1986-90 & 1987-91 & 1988-92  & 1989-93  \\
				\hline
				5 years  & 1985-90 & 1986-91 & 1987-92 & 1988-93  & 1989-94 \\
                \hline
			\end{tabular}
			\caption{Overlapped UNOS-I datasets from 1985 to 2000}
			\label{tab:UNOS-I}
		\end{small}
	\end{table*}
}

{
	\renewcommand{\arraystretch}{1.5}
	\begin{table*}[h]
		\centering
		\begin{small}
			\begin{tabular}{|c|c|c|c|c|c|}
				\hline
				 \backslashbox{Overlaps}{Years} 
                    & \multicolumn{5}{|c|}{UNOS-II}\\
				\hline
				1 year  & 1989-92 & 1992-95 & 1995-98 & 1998-01  & 2001-04 \\
				\hline
				2 years  & 1989-92 & 1991-94 & 1993-96 & 1995-98  & 1997-00 \\
				\hline
				3 years  & 1989-92 & 1990-93 & 1991-94 & 1992-95  & 1993-96 \\
				\hline
				4 years  & 1989-93 & 1990-94 & 1991-95 & 1992-96  & 1993-97 \\
				\hline
				5 years  & 1989-94 & 1990-95 & 1991-96 & 1992-97  & 1993-98 \\
				\hline
			\end{tabular}
			\caption{Overlapped UNOS-II datasets from 1989 to 2004}
			\label{tab:UNOS-II}
		\end{small}
	\end{table*}
}

{
	\renewcommand{\arraystretch}{1.5}
	\begin{table*}[h]
		\centering
		\begin{small}
			\begin{tabular}{|c|c|c|c|c|c|}
				\hline
				 \backslashbox{Overlaps}{Years} 
                    & \multicolumn{5}{|c|}{UNOS-II}\\
				\hline
				1 year  & 1989-92 & 1992-95 & 1995-98 & 1998-01  & 2001-04 \\
				\hline
				2 years  & 1989-92 & 1991-94 & 1993-96 & 1995-98  & 1997-00 \\
				\hline
				3 years  & 1989-92 & 1990-93 & 1991-94 & 1992-95  & 1993-96 \\
				\hline
				4 years  & 1989-93 & 1990-94 & 1991-95 & 1992-96  & 1993-97 \\
				\hline
				5 years  & 1989-94 & 1990-95 & 1991-96 & 1992-97  & 1993-98 \\
				\hline
			\end{tabular}
			\caption{Overlapped UNOS-III datasets from 1989 to 2004}
			\label{tab:UNOS-III}
		\end{small}
	\end{table*}
}

\clearpage
\subsection*{Computation Time of \textsc{LBO}}
\begin{figure*}[h]
	\begin{center}
		\centerline{\includegraphics[height=5.0cm,width=6cm]{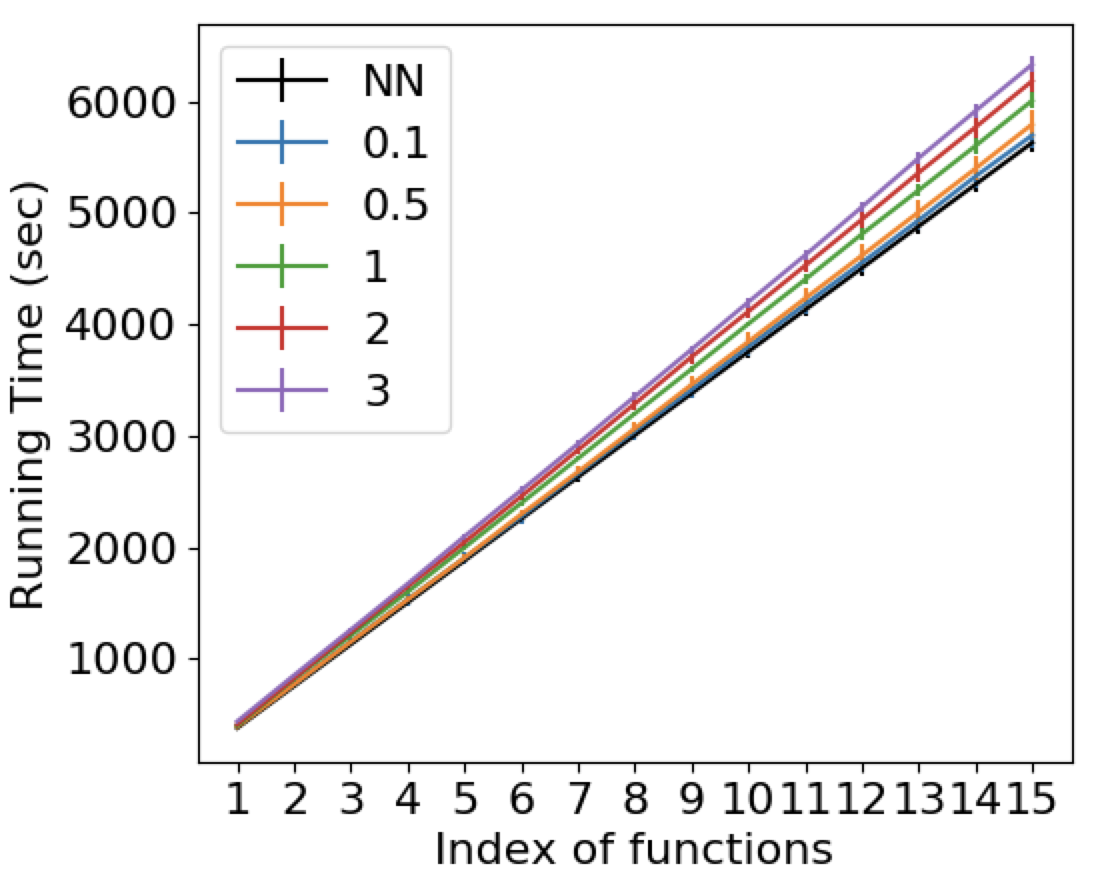}}
	\end{center}
	\caption{The accumulated running time of \textsc{LBO} with $\alpha=\{0.1, 0.5, 1, 2, 3\}$. As a comparison, the black line, denoted by \quotes{NN}, is the accumulated running time by a 3-layers neural network with 100 units at each layer, trained by marginal likelihood optimization. The neural network \quotes{NN} optimizes each black-box function independently.}
	\label{running_time}
	\vskip -0.1in
\end{figure*}

In this section, we show the accumulated computation time of \textsc{LBO} in optimizing a sequence of black-box functions, with different values of the IBP hyperparameter $\alpha$. We use \textsc{LBO} to optimize the model selection annually for the UNOS-II data from 1994 to 2008. We choose this period because the data in every year consists of samples in both classes. The AUC performance on the data of each year is treated as a black-box function. We optimize a sequence of fifteen black-box functions. In \textsc{LBO}, the truncated number of neural networks, $M$, is set to 25. Each neural network has three layers each with fifty $\tanh$ units. The experimental result is averaged over 10 runs. In Figure \ref{running_time}, we show the accumulated running time of \textsc{LBO} with $\alpha= \{0.1, 0.5, 1, 2, 3\}$. The accumulated running time only takes into account the time spent on ELBO optimization but not the process of evaluating the black-box functions with the hyperparameter selected by \textsc{LBO}. We can see the times increase linearly with an increasing number of black-box functions. Updating a large truncated number of neural networks by marginal likelihood optimization is computationally expensive since it needs to invert a $MD^{*}\times MD^{*}$ matrix at every iteration. In \textsc{LBO}, we reduce the computation to an affordable level by using an IBP prior, in which only a small number of neural networks is activated by sampling at each iteration of ELBO optimization. Therefore, \textsc{LBO} is lifelong in speeding up the hyperparameter optimization process for new datasets using the information from past datasets. 
\subsection*{Hyperparameter Space in the UNOS Experiments}
In the experiments of the UNOS datasets, we considered the model-and-hyperparameter selection problem over the algorithms XGBoost \cite{chen2016xgboost}, Logistic Regression \cite{mccullagh1989generalized} and Bernoulli naive bayes \cite{mccallum1998comparison}, Multinomial naive bayes \cite{kibriya2004multinomial}. XGBoost is implemented with the Python package \texttt{xgboost}. Other models are implemented with the \texttt{scikit-learn} library \cite{scikit-learn}. The hyperparameter space $\mathbb{X}$ is nine-dimensional.

The XGBoost model consisted of the following 3 hyperparameters:
\begin{itemize}
	\item Number of estimators (type: int, min: 10, max: 500)
	\item Maximum depth (type: int, min: 1, max: 10)
	\item Learning rate (type: float, min: 0.005, max: 0.5)
\end{itemize}

The Logistic Regression model consisted of the following 3 hyperparameters:
\begin{itemize}
	\item Inverse of regularization strength (type: float, min: 0.001, max: 10) , 
	\item Optimization solver (type: categorical, solvers: Newton-Conjugate-Gradient, Limited-memory BFGS, Lblinear, Sag, Saga)
	\item Learning rate (type: float, min: 0.005, max: 0.5)
\end{itemize}
Both Bernoulli naive bayes and Multinomial naive bayes model have one single hyperparameter:
\begin{itemize}
	\item Additive smoothing parameter (type: float, min: 0.005, max: 5)
\end{itemize}

In addition, we have a categorical variable indicating which model is selected:
\begin{itemize}
	\item Selected Model (type: categorical, algorithms: XGboost, Logistic Regression, Bernoulli naive bayes, Multinomial naive bayes)
\end{itemize}

\subsection*{UNOS-I}
\begin{figure}[h]
	\centering
	  \subfloat[]{
	\begin{minipage}{0.5\textwidth}
	   \centering
	   \includegraphics[height=4.5cm, width=5.5cm]{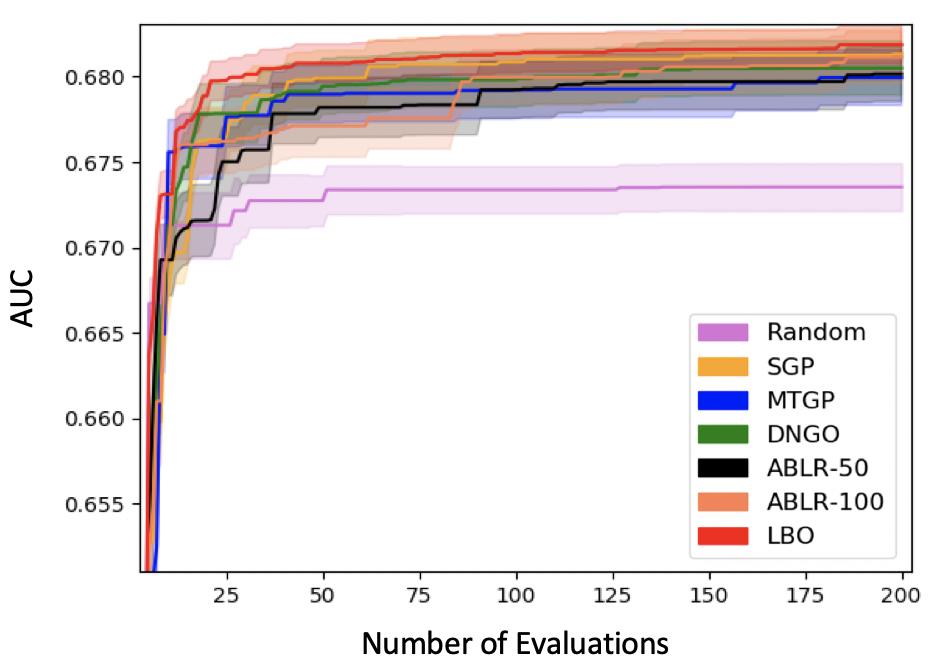}
	\end{minipage}}
	  \subfloat[]{
	\begin{minipage}{0.5\textwidth}
	   \centering
	   \includegraphics[height=4.5cm, width=5.5cm]{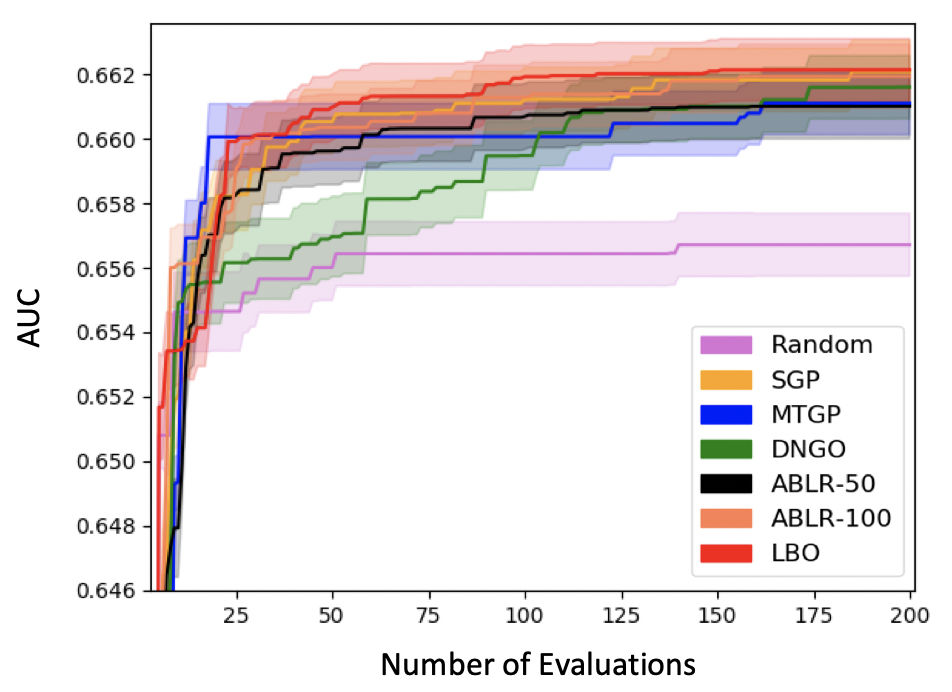}
	\end{minipage}}
	\hfill
	  \subfloat[]{
	\begin{minipage}{0.5\textwidth}
	   \centering
	   \includegraphics[height=4.5cm, width=5.5cm]{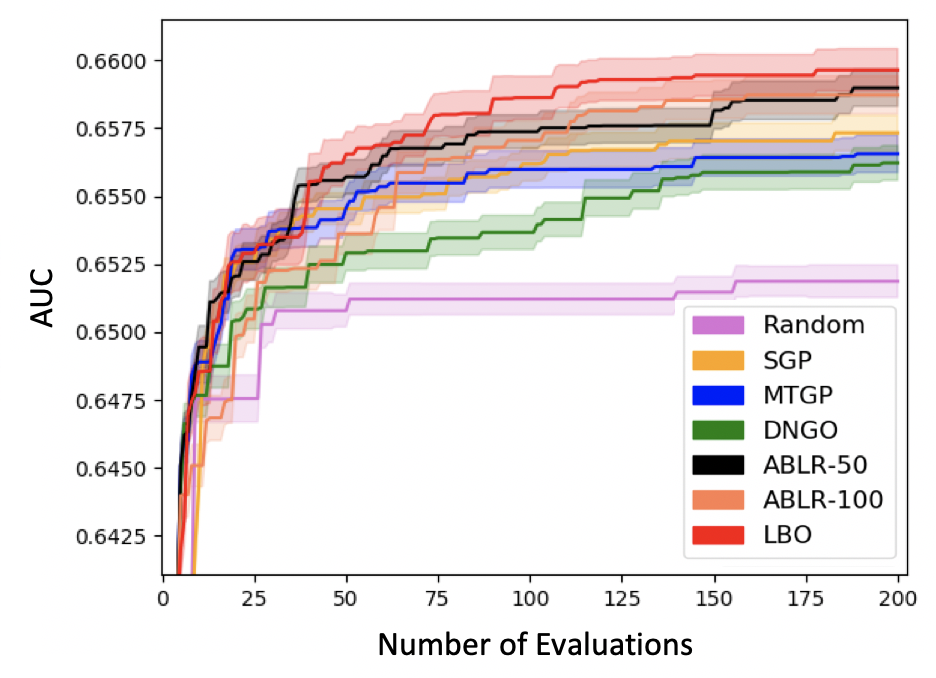}
	\end{minipage}}
	   \subfloat[]{
	\begin{minipage}{0.5\textwidth}
	   \centering
	   \includegraphics[height=4.5cm, width=5.5cm]{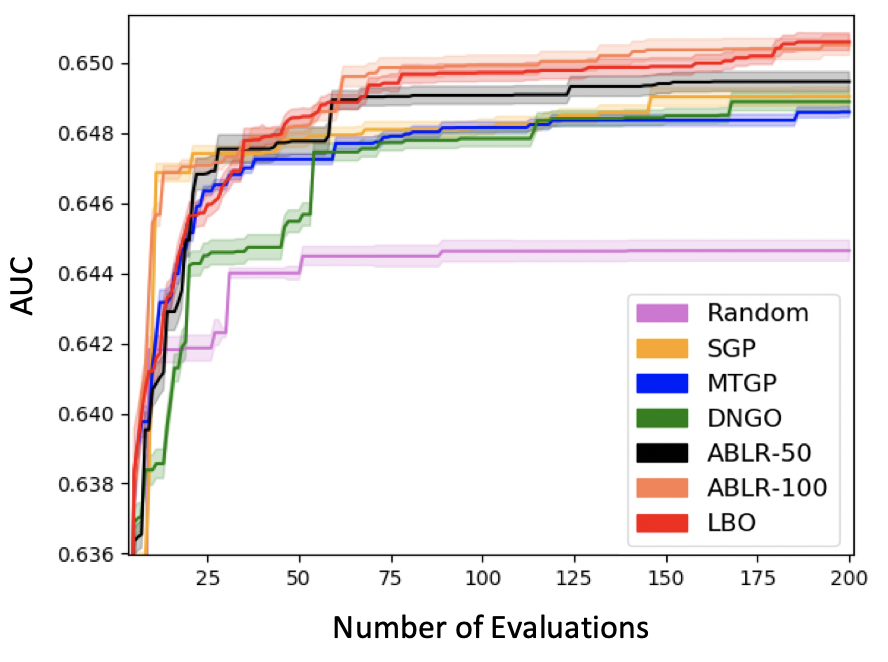}
	\end{minipage}}
	\hfill
	   \subfloat[]{
	\begin{minipage}{0.5\textwidth}
	   \centering
	   \includegraphics[height=4.5cm, width=5.5cm]{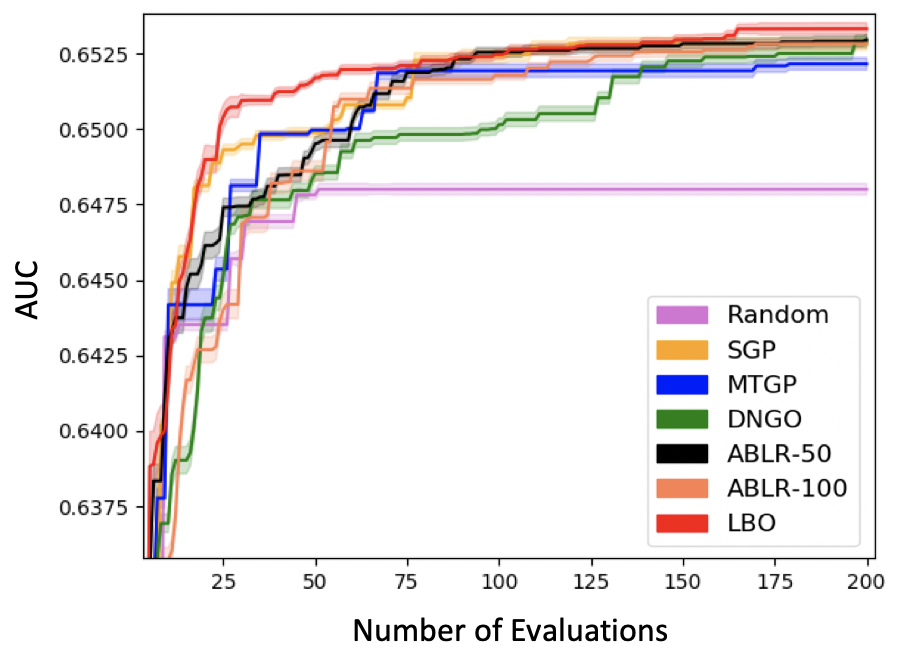}
	\end{minipage}}
\caption{200 function evaluations of BO algorithms on UNOS-I: (a) 1-year Overlap; (b) 2-years Overlap; (c) 3-years Overlap; (d) 4-years Overlap;  (e) 5-years Overlap.}
\vskip -0.1in
\end{figure}

\clearpage
\newpage
\subsection*{UNOS-II}
\begin{figure}[h]
	\centering
	  \subfloat[]{
	\begin{minipage}{0.5\textwidth}
	   \centering
	   \includegraphics[height=4.5cm, width=5.5cm]{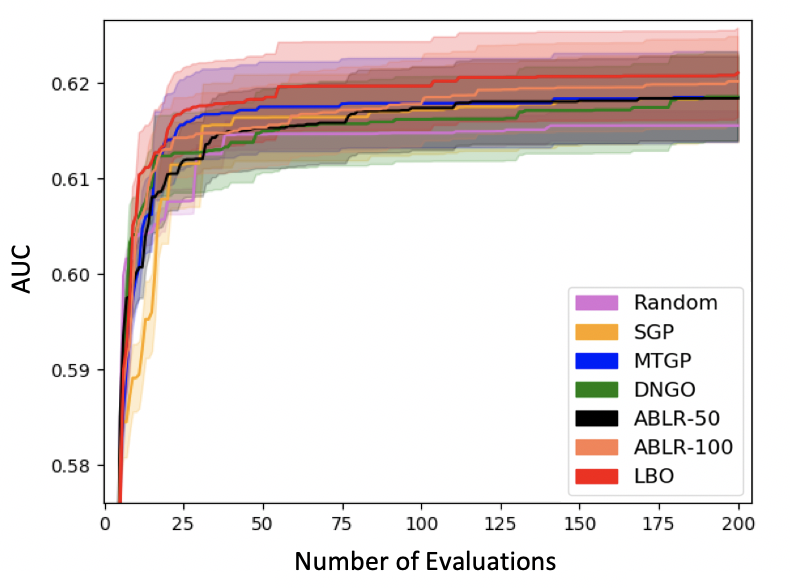}
	\end{minipage}}
	  \subfloat[]{
	\begin{minipage}{0.5\textwidth}
	   \centering
	   \includegraphics[height=4.5cm, width=5.5cm]{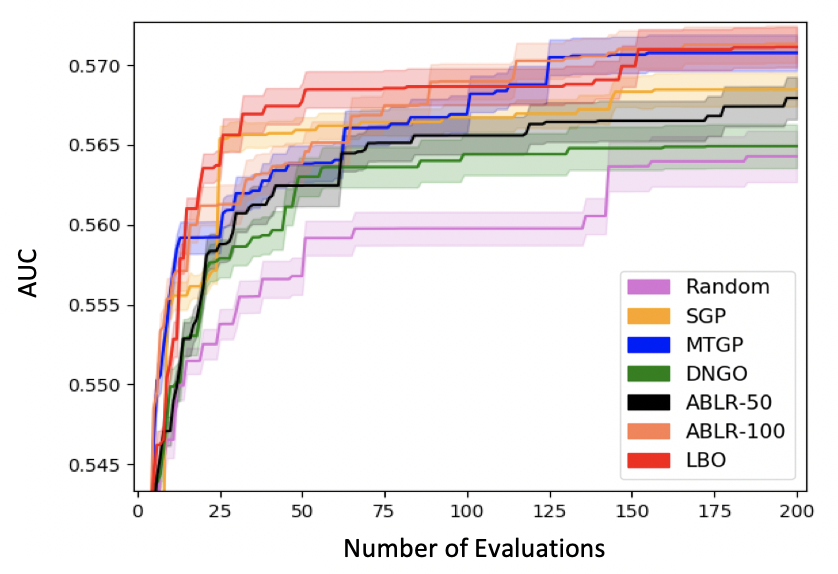}
	\end{minipage}}
	\hfill
	  \subfloat[]{
	\begin{minipage}{0.5\textwidth}
	   \centering
	   \includegraphics[height=4.5cm, width=5.5cm]{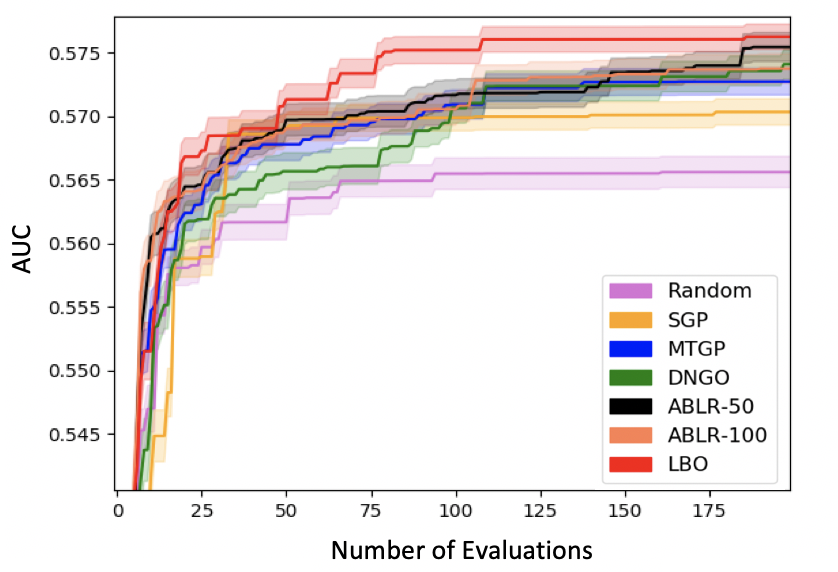}
	\end{minipage}}
	   \subfloat[]{
	\begin{minipage}{0.5\textwidth}
	   \centering
	   \includegraphics[height=4.5cm, width=5.5cm]{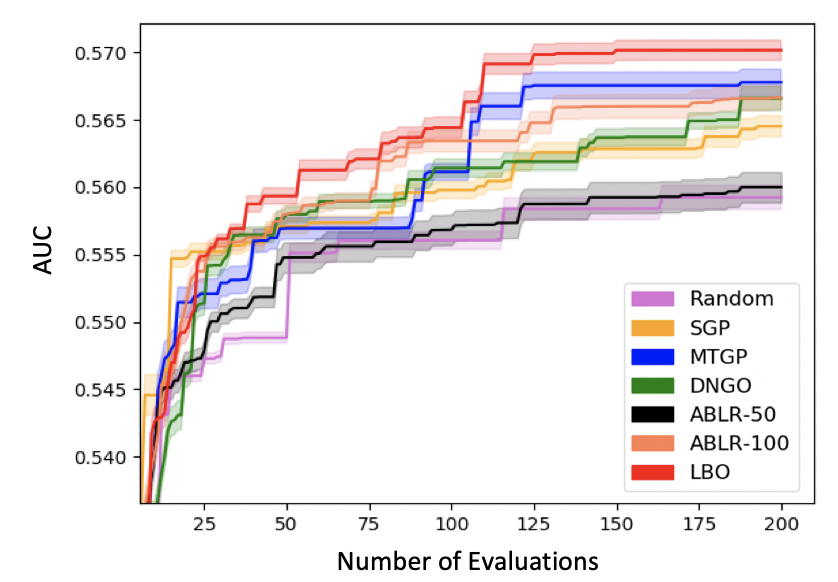}
	\end{minipage}}
	\hfill
	   \subfloat[]{
	\begin{minipage}{0.5\textwidth}
	   \centering
	   \includegraphics[height=4.5cm, width=5.5cm]{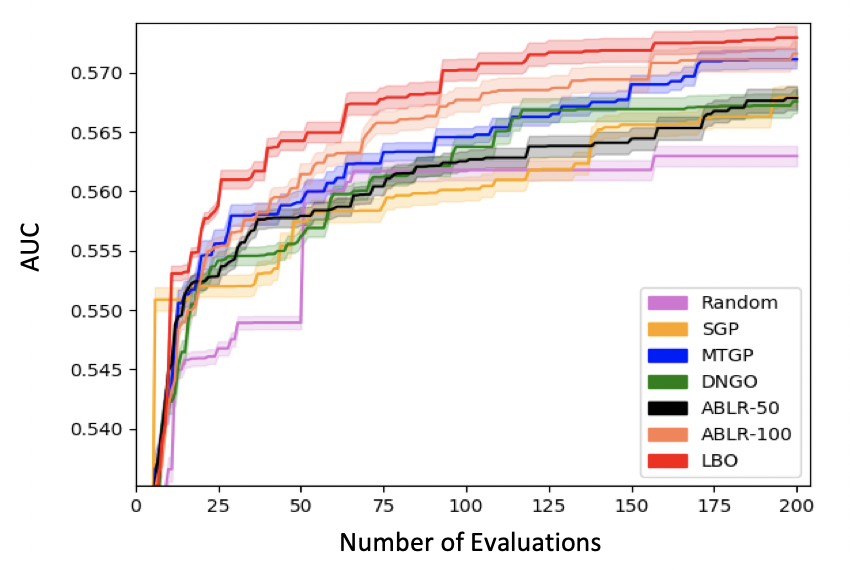}
	\end{minipage}}
\caption{200 function evaluations of BO algorithms on UNOS-II: (a) 1-year Overlap; (b) 2-years Overlap; (c) 3-years Overlap; (d) 4-years Overlap;  (e) 5-years Overlap.}
\vskip -0.1in
\end{figure}

\clearpage
\newpage
\subsection*{UNOS-III}
\begin{figure}[h]
	\centering
	  \subfloat[]{
	\begin{minipage}{0.5\textwidth}
	   \centering
	   \includegraphics[height=4.5cm, width=5.5cm]{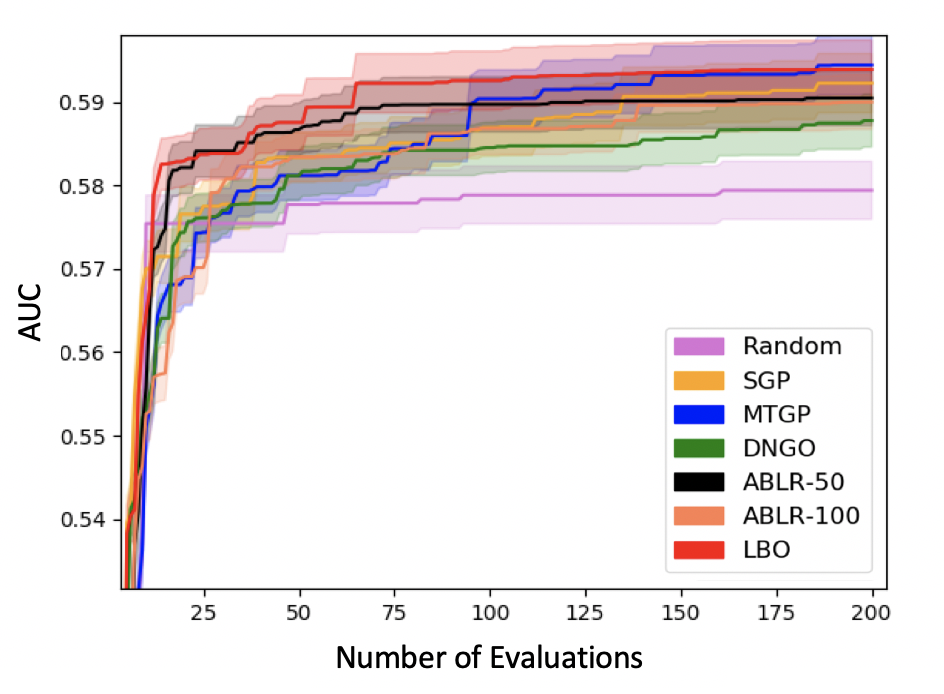}
	\end{minipage}}
	  \subfloat[]{
	\begin{minipage}{0.5\textwidth}
	   \centering
	   \includegraphics[height=4.5cm, width=5.5cm]{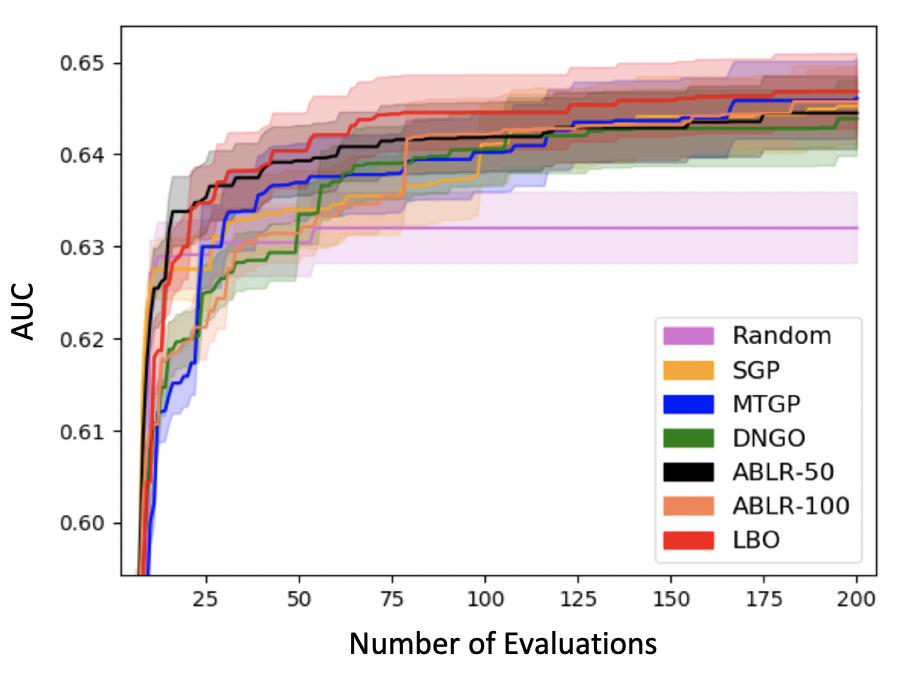}
	\end{minipage}}
	\hfill
	  \subfloat[]{
	\begin{minipage}{0.5\textwidth}
	   \centering
	   \includegraphics[height=4.5cm, width=5.5cm]{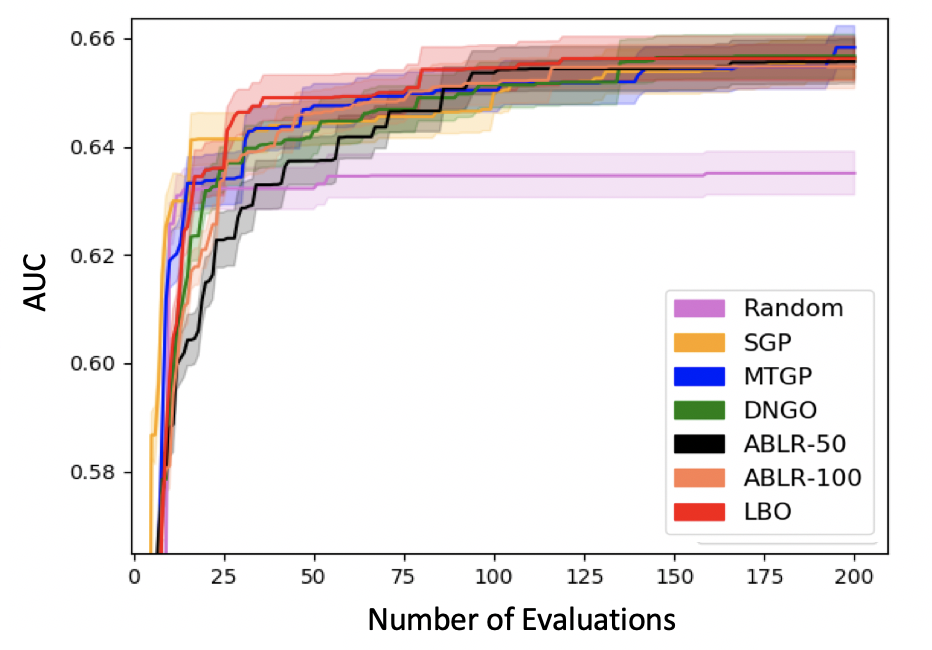}
	\end{minipage}}
	   \subfloat[]{
	\begin{minipage}{0.5\textwidth}
	   \centering
	   \includegraphics[height=4.5cm,  width=5.5cm]{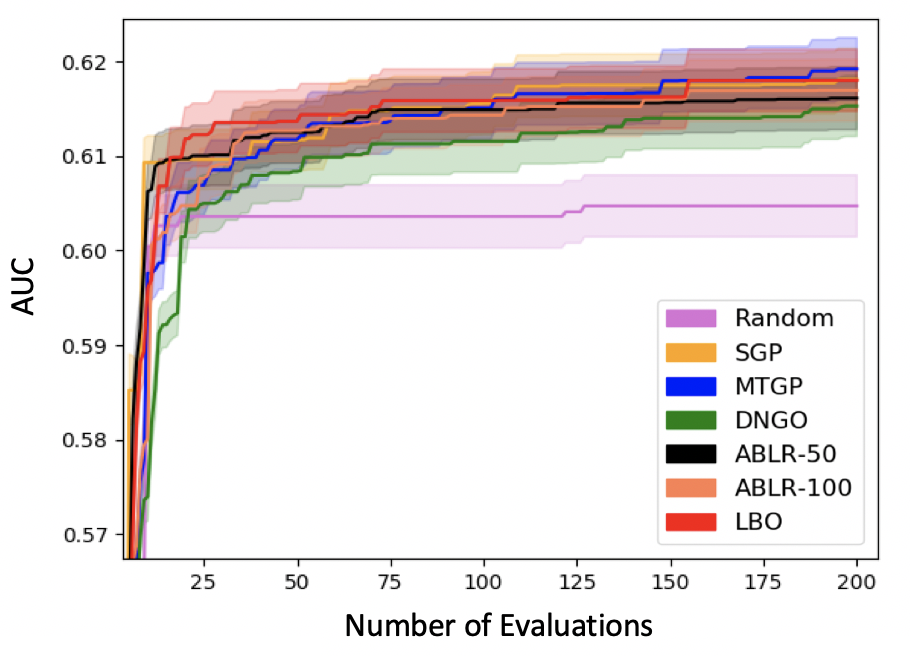}
	\end{minipage}}
	\hfill
	   \subfloat[]{
	\begin{minipage}{0.5\textwidth}
	   \centering
	   \includegraphics[height=5cm, width=6.2cm]{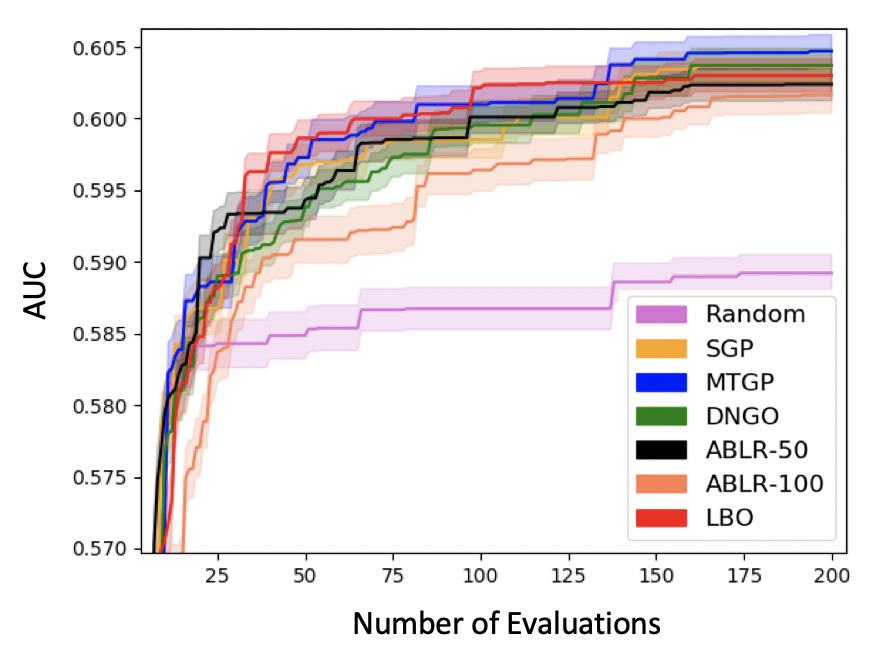}
	\end{minipage}}
\caption{200 function evaluations of BO algorithms on UNOS-III: (a) 1-year Overlap; (b) 2-years Overlap; (c) 3-years Overlap; (d) 4-years Overlap;  (e) 5-years Overlap.}
\vskip -0.1in
\end{figure}

\clearpage
\newpage

\section*{Pseudo-code for Lifelong Bayesian Optimization (\textsc{LBO})}
\begin{algorithm}[htbp]
	\caption{Lifelong Bayesian Optimization}
	\label{alg:main}
	\begin{algorithmic}
		\STATE {\bfseries Hyperparameters:} Maximum number of acquisitions $N_{1}$, Maximum number of training iterations $N_{2}$, Indian Buffet Process parameter $\alpha$, Truncated number of neural network $M$
		\STATE {\bfseries Input:} New dataset $\mathcal{D}_t$ (or equivalently new black-box function $f_t$), 
		acquisition function $a_{f_t}$ 
		\STATE {\bfseries Initialize:} $\mathcal{A}_t$ by performing some random acquisitions of $f_t$ 
		\FOR{$i=1$ {\bfseries to} $N_{1}$}
		\FOR{$j=1$ {\bfseries to} $N_{2}$}
		\STATE Sampling  $\pi_m$ $z_{t,m}$ using (\ref{equ:IBP}) and (\ref{equ:concrete})
		\STATE Update $\mathbf{\Theta}_t$ and  $\{\gamma_{t,m}\}_{m=1}^{M}$ by optimizing the ELBO in (\ref{equ:ELBO}) regularized by (\ref{eq:reg})
		\ENDFOR
		\STATE Update $a_{f_t}(\bm{x})$ using (\ref{equ:mean_var})
		\STATE Solve $\bm{x}^*=\argmax_{\bm{x}\in \mathbb{X}}a_{f_t}(\bm{x})$,
		\STATE $\mathcal{A}_{t}\leftarrow \mathcal{A}_t \cup (\bm{x}^{*},y^{*}) $
		\ENDFOR
		\STATE {\bfseries Output:} $(\bm{x},y)$ with the maximum value of $y$ in $\mathcal{A}_{t}$
	\end{algorithmic}
\end{algorithm}

\section*{Table of mathematical notations}

\begin{table}[H]
	\vskip 0.15in
	\begin{center}
		\begin{small}
			\begin{tabular}{lcr}
				\toprule
				Notation & Definition\\
			    $D$  & The input dimension of the hyperparameter space $\mathbb{X}$\\
				$D^{*}$  & The dimension of the feature map $\bm{\phi}_{t,m}$  \\
				$f_t$ & The $t$th black-box function\\ 
				$s_{t,m}$   & The weight of $m$th latent function in $f_t$\\
				$z_{t,m}$   & Indicator of whether $g_m$ is used in $f_t$\\
				$g_m$    & The $m$th latent function (neural network)\\
				$\bm{\phi}_{t,m}$  & the $m$th feature map in the spanning of $f_t$ \\
				$\mathbf{W}_{t,m}$ & the parameters in the feature map $\bm{\phi}_{t,m}$\\
				$\mathbf{\Theta}_t$ & the parameters of all the feature maps $\bm{\phi}_{t,m}$, $m=1,...,M$\\
				$\mathbf{h}_m$  & The output layer of the $m$th neural network \\
				$a_{f}$   & Acquisition function \\
				$\mathcal{A}_t$ & Acquisition set for $f_t$\\
				$\mathcal{A}_{\leq t}$  & The union of acquisition sets up to time $t$ \\
				$\mathbf{X}_t$&  Input data in $\mathcal{A}_t$  \\
				$\mathbf{Y}_t$&  Output data in $\mathcal{A}_t$ \\
				$N_t$  & The total number of acquisitions of $f_t$ \\
				$M$ & The truncated number of latent functions in IBP\\
				$\alpha$ & the parameter of Indian Buffet Process\\
				$v_m$ & the $m$th Beta sample in the Stick breaking Process \\
				$\pi_m$ & the probability of using the $m$th neural network \\
				$\gamma_{t,m}$ & the variational parameter in the $\BinConcrete$ distribution
				$Q_{\gamma_{t,m},\tau_{t,m}}$\\
				$\tau_{t,m}$ & the temperature hyperparameter in the $\BinConcrete$ distribution
				$Q_{\gamma_{t,m},\tau_{t,m}}$\\
				$\lambda_t$ & precision parameter in the prior distribution $P(\mathbf{h}_{t,m}|\lambda_t)$\\
				$\beta_t$ & precision parameter in the likelihood distribution $P(\mathbf{Y}_t|\mathbf{f}_t,\beta_t)$ \\
				\bottomrule
			\end{tabular}
			\vskip 0.15in
			\caption{Table of mathematical notations}
		\end{small}
	\end{center}
	\vskip -0.1in
\end{table}

\end{document}